\documentclass{article}

\usepackage[preprint]{neurips_2026}

\usepackage[utf8]{inputenc} 
\usepackage[T1]{fontenc}    
\usepackage{hyperref}       
\usepackage[acronym, nogroupskip, nonumberlist]{glossaries-extra}
\setkeys{glslink}{hyper=false}
\makenoidxglossaries
\setabbreviationstyle[acronym]{long-short} 

\newabbreviation{mi}{MI}{mutual information}
\newabbreviation{kl}{KL}{Kullback-Leibler}
\newabbreviation{tdmi}{TDMI}{time-delayed mutual information}
\newabbreviation{te}{TE}{transfer entropy}
\newabbreviation{ais}{AIS}{active information storage}
\newabbreviation{cd}{CD}{causal density}
\newabbreviation{ii}{II}{integrated information}

\newabbreviation{pid}{PID}{Partial Information Decomposition}
\newabbreviation{phiid}{$\Phi$ID}{Integrated Information Decomposition}

\newabbreviation{mmi}{MMI}{minimum mutual information}
\newabbreviation{ccs}{CCS}{common change in surprisal}

\newabbreviation{poset}{poset}{partially ordered set}

\newabbreviation{sde}{SDE}{stochastic differential equation}
\newabbreviation{vpsde}{VP-SDE}{variance-preserving stochastic differential equation}
\newabbreviation{ema}{EMA}{exponential moving average}
\newcommand{\E}[2]{\mathbb{E}_{#1 \sim #2}}
\newcommand{\KL}[2]{D_{\text{KL}} \left[ #1 \ \vert \vert \ #2 \right]}
\newcommand{\Score}[2]{\displaystyle S^{#1_{#2}}}

\newcommand{\Eqref}[1]{\hyperref[#1]{Eq.~(\ref*{#1})}}
\newcommand{\Secref}[1]{\hyperref[#1]{\S~\ref*{#1}}}
\newcommand{\Figref}[1]{\hyperref[#1]{Figure~\ref*{#1}}}
\newcommand{\Algoref}[1]{\hyperref[#1]{Algorithm~\ref*{#1}}}
\usepackage{url}            
\usepackage{booktabs}       
\usepackage{amsfonts}       
\usepackage{nicefrac}       
\usepackage{microtype}      
\usepackage{xcolor}         
\usepackage{graphicx}       
\usepackage{amsthm}
\newtheorem{theorem}{Theorem}
\newtheorem{assumption}{Assumption}
\usepackage{dsfont}
\usepackage{ulem}
\usepackage{float}
\usepackage{tikz}
\usetikzlibrary{positioning, calc}
\definecolor{redcol}{RGB}{240,150,140}
\definecolor{yellcol}{RGB}{245,220,130}
\definecolor{bluecol}{RGB}{160,195,230}
\definecolor{orangecol}{RGB}{245,185,110}

\newcommand{\hidecomment}[1]{}

\title{DIPHINE: Diffusion-based $\Phi$-ID Neural Estimator}

\author{
  Simon Pedro Galeano Mu\~{n}oz \\ KAUST, Saudi Arabia \And
  Mustapha Bounoua \\EURECOM, France \And  
    Giulio Franzese \\EURECOM, France \AND
    Pietro Michiardi \\EURECOM, France \And
    Maurizio Filippone \\ KAUST, Saudi Arabia
}

\begin{document}

\maketitle

\begin{abstract}
Uncovering the true informational architecture of real-world complex systems requires disentangling how their components uniquely store, redundantly share, and synergistically integrate information over time. 
Integrated Information Decomposition ($\Phi$ID) is a framework for decomposing the information dynamics of multivariate systems into sixteen non-overlapping atoms that characterize redundant, unique, and synergistic modes of information storage, transfer, and integration. 
Existing methods to compute $\Phi$ID are restricted to Gaussian or discrete systems, preventing its application to continuous non-Gaussian dynamical systems. We address this limitation by proposing DIPHINE (Diffusion-based $\Phi$-ID Neural Estimator), the first neural estimator that leverages score-based diffusion models to jointly estimate all the mutual information terms required by $\Phi$ID from a single amortized network, recovering the sixteen atoms through M\"obius inversion. We provide a theoretical analysis of error propagation through the inversion, showing that the Jacobian of the mapping from mutual informations to atoms is integer-valued and that the synergy-to-synergy atom is provably the hardest to estimate. We demonstrate accurate recovery of ground-truth atoms on synthetic benchmarks, superior performance compared to established mutual information estimators, and the ability to extract physiologically interpretable information-dynamic structure on an application involving real data without any distributional assumptions.
\end{abstract}

\section{Introduction}
\label{sec:introduction}


Many natural phenomena are driven by complex, continuous interactions that cannot be fully captured by traditional, aggregated summary statistics. Understanding how the components of a complex system store, transfer, and integrate information is a central challenge in fields ranging from neuroscience \citep{luppi2024information} to physiology \citep{iyengar1996age} and complex systems analysis \citep{mediano2025toward}. For a dynamical system $X_t = [X_{1, t}, \ X_{2, t}]$, classical measures like transfer entropy \citep{schreiber2000measuring} provide scalar summaries of directed flow, but they conflate qualitatively different modes of interaction; specifically, whether information is redundantly shared, uniquely carried, or synergistically integrated.


\Gls{phiid}, introduced by \citet{mediano2025toward}, addresses this limitation by decomposing the total information flow between $X_t$ and $X_{t+1}$ into sixteen non-overlapping components called atoms. These atoms form a complete taxonomy from which well-known measures of information can be recovered. 
While \gls{phiid} reveals fine-grained structure invisible to conventional summaries \citep{luppi2024synergistic}, its practical use is currently confined to Gaussian \citep{barrett2015exploration} or discrete systems \citep{ince2017measuring}. For continuous non-Gaussian systems, no general-purpose estimator exists.

We address this gap by proposing DIPHINE (Diffusion-based $\Phi$-ID Neural Estimator). In summary, our main contributions are twofold: \textbf{(1) A Novel Neural Estimator:} We introduce the first method capable of estimating all sixteen \gls{phiid} atoms from continuous data without distributional assumptions, leveraging a single amortized score-based diffusion network \citep{song2020score, franzese2023minde} to compute the nine necessary \gls{mi} terms; and \textbf{(2) Theoretical Error Propagation Analysis:} We prove that under reasonable assumptions the Jacobian mapping from estimated \gls{mi}s to \gls{phiid} atoms has a closed form, furthermore, we show that it is integer-valued, providing theoretical bounds that characterize the synergy-to-synergy atom as strictly the hardest to estimate. Finally, we validate DIPHINE on synthetic benchmarks and demonstrate its ability to extract physiologically interpretable information-dynamic structure from real-world cardio-respiratory data.
\section{Background}
\subsection{Mutual Information}
\label{sec:MI}
Measuring dependencies between random variables arises frequently in several applications of Statistics and Machine learning \citep{Vicenteetal2011, KaoGovindaraju2010, Patton2006}. Even though there exist several proposals for such regard \citep{Szekelyetal2008, Yuetal2021, Grettonetal2005}, \Gls{mi} stands out as it does not require any assumption on functional relationships between variables (e.g., linear, monotonic, etc) as it is agnostic to such relationships. Let $X \in \mathbb{R}^{d_x}$ and $Y \in \mathbb{R}^{d_y}$ be random variables with joint density $p_{X, Y}$ and marginal densities $p_X$ and $p_Y$ respectively. The \gls{mi} between $X$ and $Y$ is defined as:\footnote{MI can be defined also for variables without densities
and in more generic spaces, but for the purpose of this work
the restriction considered here is sufficient.}
\begin{equation}
    \label{eq:MI}
    I(X; Y) = \KL{p_{X, Y}}{p_X \, p_Y},
\end{equation}
where $\KL{p}{q} = \E{x}{p}\left[\log\left(\frac{p(x)}{q(x)}\right)\right]$ denotes the \gls{kl} divergence \citep{KullbackLeibler1951}. It is worth noticing that  $I(X; Y) = 0$ if and only if $X$ and $Y$ are independent, moreover, it is a symmetric measure of dependence, i.e. $I(X; Y) = I(Y; X)$. 

Throughout this work, we consider $d$-dimensional discrete time ergodic processes $\left \{X_t \right\}$\footnote{So that all quantities of interest are independent of $t$ and can be estimated from a single long realization of the process.} that are disjointly partitioned as $X_t = \left[ X_{1, t}, \ X_{2, t} \right]$ with $X_{1, t} \in \mathbb{R}^{d_1}$ and $X_{2, t} \in \mathbb{R}^{d_2}$, such that $d_1 + d_2 = d$. To study information dynamics, we examine how the current state $X_t$ of the system relates to its future state $X_{t+1}$. The \gls{tdmi} captures the total information that the present conveys about the future of the system:
\begin{equation}
    \label{eq:TDMI}
    \text{TDMI} = I(X_t; X_{t+1}) = I(X_{1, t}, \ X_{2, t}; \ X_{1, t+1}, \ X_{2, t+1}).
\end{equation}
While \gls{tdmi} provides a single scalar summary of the dependence between the current state of the system and its dependency with the future state, it qualitatively merges different modes of information flow between components of the present and future, such as information that is shared redundantly across variables, carried uniquely by one of them, or available only when both are considered jointly. The framework described in the following subsections address this limitation by decomposing the \gls{tdmi} into fine-grained components. 
\subsection{Partial Information Decomposition}
\label{sec:PID}
The \Gls{pid} framework, introduced by \citet{williams2010nonnegative}, decomposes the information that a collection of source variables conveys about a single target variable. Consider two source variables $X_1, X_2$ and a target variable $Y$. The joint \gls{mi} $I(X_1, X_2; Y)$ can be expressed as the sum of four quantities:
\begin{equation}
    \label{eq:PID}
    I(X_1, X_2; Y) = \underbrace{R(X_1, X_2 \to Y)}_{\text{redundancy}} + \underbrace{U(X_1 \to Y)}_{\text{unique}_1} + \underbrace{U(X_2 \to Y)}_{\text{unique}_2} + \underbrace{S(X_1, X_2 \to Y)}_{\text{synergy}},
\end{equation}
where the redundancy $R$ captures the information about $Y$ that is available from either source alone. The unique terms $U(X_i \to Y)$ capture the information available exclusively from $X_i$, and the synergy $S$ captures the information that becomes accessible only when both sources are observed jointly.

These four quantities correspond to the elements of a lattice structure (the redundancy lattice) that organizes all qualitatively distinct ways in which the sources can provide information about the target\footnote{The construction of the redundancy lattice and a detailed discussion of the anti-chains structure are provided in \Secref{app:lattice_construction}.}. 

\subsubsection{Lattices, M\"obius Inversion and Atoms}
\label{sec:posets}
The mathematical structure underlying information decomposition is rooted in the theory of \Glspl{poset} \cite{liu2024information}. A \gls{poset} is a pair $(\mathcal{A}, \preceq)$ consisting of a set $\mathcal{A}$ together with a relation $\preceq$ that is reflexive, antisymmetric, and transitive\footnote{It is possible that for some $\alpha, \beta \in \mathcal{A}$ neither $\alpha \preceq \beta$ nor $\beta \preceq \alpha$ holds.}. A \gls{poset} $(\mathcal{A}, \preceq)$ is called a lattice if every pair of elements $\alpha, \beta \in \mathcal{A}$ admits a greatest lower bound and a least upper bound. 

Given a finite \gls{poset} $(\mathcal{A}, \preceq)$, the M\"obius function $\mu: \mathcal{A} \times \mathcal{A} \to \mathbb{Z}$ is defined as follows: 
\begin{equation}
    \label{eq:mobius_function}
    \mu(\alpha, \beta) = \mathds{1}_{\{\alpha = \beta\}} - \mathds{1}_{\{\alpha \prec \beta\}} \left(\sum_{\alpha \preceq \gamma \prec \beta} \mu(\alpha, \gamma) \right),
\end{equation}
where $\mathds{1}_{\{\cdot\}}$ is the indicator function and for any two elements $\alpha, \beta \in \mathcal{A}$, $\alpha \prec \beta$ denotes $\alpha \preceq \beta$ but $\alpha \neq \beta$. 
The M\"obius function enables the inversion of cumulative relationships on a lattice, that is, if $f, g: \mathcal{A} \to \mathbb{R}$ satisfy $g(\beta) = \sum_{\alpha \preceq \beta} f(\alpha)$ for all $\beta \in \mathcal{A}$, then $f$ can be recovered via M\"obius inversion as
\begin{equation}
    \label{eq:mobius_inversion}
    f(\alpha) = \sum_{\beta \preceq \alpha} \mu(\beta, \alpha) \, g(\beta).
\end{equation}
Intuitively, the M\"obius function provides the correction coefficients required to undo cumulative sums over the order structure of the \gls{poset}, akin to a deconvolution process. In the same spirit as the inclusion–exclusion principle, it compensates for repeated counting induced by overlaps among lower elements. 

A cumulative redundancy function $I_\cap(\alpha, \beta)$ is defined on this lattice, where each node represents a cumulative quantity of information shared across all lower-order source collections. M\"obius inversion then removes these nested overlaps by correcting for accumulated information, thereby isolating the irreducible \gls{pid} atoms corresponding to uniquely attributable redundant, unique, and synergistic contributions. Such atomic quantities are recovered via M\"obius inversion of $I_\cap$, as described in \Eqref{eq:mobius_inversion}. Different choices of $I_\cap$ lead to different decompositions; in this work, we adopt the \gls{mmi} redundancy \citep{barrett2015exploration}, defined as $I_\cap^\text{MMI}(\alpha; Y) = \min_{\mathbf{a} \in \alpha} I(\mathbf{a}; Y)$, noting that it is the only choice for which a closed-form solution exists in the Gaussian case.
\subsection{Integrated Information Decomposition}
\label{sec:phiid}
While \gls{pid} decomposes the information that multiple sources convey about a single target, many applications in complex systems involve multi-target settings where the goal is to understand how the present state of a system informs its future. \Gls{phiid}, introduced by \citet{mediano2025toward}, extends the \gls{pid} framework to handle multiple targets simultaneously, enabling a complete decomposition of the information dynamics of multivariate systems.
 
Consider the setting introduced in \Secref{sec:MI} with sources $\left \{ X_{1,t}, X_{2,t} \right \}$ and targets $\left \{ X_{1,t+1}, X_{2,t+1} \right \}$. Both the source side and the target side admit the same four-element redundancy lattice described in \Secref{sec:PID}. The \gls{phiid} framework constructs a double redundancy lattice as the Cartesian product $\mathcal{A}_\Phi := \mathcal{A}_s \times \mathcal{A}_t$ of the source and target lattices respectively, with the partial order defined componentwise: 

\begin{equation*}
(\alpha, \beta) \preceq (\alpha', \beta') \iff \alpha \preceq \alpha' \text{ and } \beta \preceq \beta'; \ \alpha, \ \alpha' \in \mathcal{A}_s, \ \beta, \ \beta' \in \mathcal{A}_t.     
\end{equation*}

The product lattice contains $4 \times 4 = 16$ elements, each corresponding to a distinct \gls{phiid} atom that characterizes a specific mode of information dynamics.
 
Each atom $(\alpha, \beta) \in \mathcal{A}_\Phi$ describes a mode of information flow from source collection $\alpha$ to target collection $\beta$. These sixteen atoms can be labeled by pairs of information types; redundancy (Red), unique to variable 1 (Un$_1$), unique to variable 2 (Un$_2$), and synergy (Syn), for both sources and targets. We denote the atom value at $(\alpha, \beta)$ by $\pi(\alpha, \beta)$. For example, $\pi(\text{Red}, \text{Syn})$ quantifies the information that is redundantly available from either variable in the present, that can only be decoded by observing both future variables jointly.
 
To compute the sixteen atoms, \gls{phiid} proceeds in two similar steps to \gls{pid}. First, a (double) redundancy function $I_\cap^{\Phi}$ is defined in the product lattice. Second, the sixteen atoms are obtained through M\"obius inversion. Using the extension\footnote{It is worth noticing that such extension may yield negative atoms as it is not totally monotonic in the double-redundancy lattice \citep{mediano2025toward}.} of \gls{mmi} redundancy to this setup \citep{mediano2025toward}, the redundancy function takes the form
\begin{equation}
    \label{eq:double_redundancy}
    I_\cap^{\Phi}(\alpha, \beta) = \min_{\mathbf{a} \in \alpha} \min_{\mathbf{b} \in \beta} I(\mathbf{a}; \mathbf{b}),
\end{equation}
where $I(\mathbf{a}; \mathbf{b})$ denotes the \gls{mi} between the concatenation of variables indexed by $\mathbf{a}$ at time $t$ and the concatenation of variables indexed by $\mathbf{b}$ at time $t + 1$. 
The sixteen \gls{phiid} atoms provide a complete decomposition of the \gls{tdmi}:
\begin{equation}
    \label{eq:phiid_sum}
    I(X_{t}; X_{t+1}) = \sum_{(\alpha, \beta) \in \mathcal{A}_\Phi} \pi(\alpha, \beta).
\end{equation}
Moreover, the \gls{phiid} framework subsumes several classical measures of information dynamics as specific aggregates of its atoms, for instance \gls{te} \citep{schreiber2000measuring}, \gls{ais} \citep{Lizier2012}, \gls{cd} \citep{Seth2005}, and \gls{ii} \citep{oizumi2016unified} can be recovered from the \gls{phiid} atoms. In this way, \gls{phiid} provides a unified taxonomy that exposes distinct modes of information dynamics that are invisible to conventional analyses of well-known information-theoretic measures.
\section{Related work}
\label{sec:related}
\paragraph{\gls{phiid} and \gls{pid}.} The \gls{pid} framework of \citet{williams2010nonnegative} has spawned a rich literature on redundancy functions, including the \gls{mmi} approach \citep{barrett2015exploration}, the maximum-entropy formulation of \citet{bertschinger2014quantifying}, and the \gls{ccs} measure of \citet{ince2017measuring}. \Gls{phiid} \citep{mediano2025toward} extends \gls{pid} to multi-target settings, enabling a complete decomposition of temporal information dynamics. However, existing methods for computing \gls{phiid} on continuous systems are restricted to the assumption of Gaussian, where the \gls{mmi} redundancy admits a closed form \citep{barrett2015exploration}. Thus, no general-purpose estimator exists. On the \gls{pid} side, recent work has addressed estimation beyond the discrete and Gaussian cases using normalizing flows \citep{zhao2025flowpid} and copula-based methods \citep{pakman2021estimating}, but these target the single-target \gls{pid} and do not extend to the multi-target \gls{phiid} setting.

\paragraph{\gls{mi} estimators.} Variational approaches such as MINE \citep{belghazi2018mine} and NWJ \citep{nguyen2010nwj} estimate \gls{mi} via bounds on the \gls{kl} divergence, but can require exponentially large datasets for tight bounds \citep{mcallester2020formal}. InfoNCE \citep{oord2018infonce} provides a more stable bound but saturates at $\log N$. Non-parametric methods such as KSG \citep{kraskov2004ksg} are effective in low dimensions but degrade with the curse of dimensionality. More recent generative approaches estimate \gls{mi} through normalizing flows \citep{butakov2024mienf} and Schr\"odinger bridges \citep{kholkin2025infobridge}, achieving improved accuracy in high dimensions. All of these methods estimate a single \gls{mi}; applying them to \gls{phiid} requires training nine separate models, without shared representations across the score functions.

\paragraph{Score-based information estimation.} Score-based diffusion models \citep{song2020score} have recently been leveraged for information-theoretic estimation. MINDE \citep{franzese2023minde} uses the Girsanov theorem to estimate \gls{kl} divergences from score differences, yielding accurate \gls{mi} estimates. S$\Omega$I \citep{bounoua2024s} extends this to estimate O-information from a single amortized score network via conditional masking. TENDE \citep{galeano2026tende} further adapts this framework to estimate transfer entropy through conditional mutual information. DIPHINE builds on this line of work, adapting the amortized masking scheme to the nine \gls{mi} structure required by \gls{phiid} and providing, for the first time, a neural estimator of all sixteen \gls{phiid} atoms.
\section{Methods}
\label{sec:methods}
\subsection{Score-based KL divergence estimation}
\label{sec:overview_scores}
Recall that $X \in \mathbb{R}^{d_x}$ denotes a random variable with density $p_X$. Under certain regularity conditions, \citet{hyvarinen2005estimation} showed that it is possible to associate the density $p_X$ with the score function $\Score{p}{X}$, where for a generic distribution $p_X$ we denote $\Score{p}{X}(x) := \nabla \log \left( p_X(x) \right)$, with derivatives taken with respect to $x$. With this in mind, it is possible to construct a diffusion process $\{X_\tau\}_{\tau \in [0, T]}$\footnote{We use $\tau$ for the diffusion time to distinguish it from the discrete process time $t$ introduced in \Secref{sec:MI}.} such that $X_0 \sim p_X$ and $X_T \sim p_{X_T}$ where $p_{X_T}$ is a known and tractable distribution. The constructed process is modeled as the solution of the following \gls{sde}:
\begin{equation}
    \label{eq:diffusion}
    \begin{cases}
        \begin{aligned}
            dX_\tau &= f_\tau X_\tau \, d\tau + g_\tau \, dW_\tau \\
            X_0 &\sim p_X,
        \end{aligned}
    \end{cases}
\end{equation}
with given continuous functions $f_\tau \leq 0, \ g_\tau \geq 0$ for each $\tau \in [0, T]$; $dW_\tau$ is a Brownian motion. The random variable $X_\tau$ is associated with its density $p_{X_\tau}$ and therefore with the time-varying score $\Score{p}{X_\tau}(x)$. In this work, we employ the \gls{vpsde} as described in \citet{song2020score} to construct the diffusion process. A key practical advantage of this formulation is that the probability density $p_{0\tau}(\cdot \mid x)$ of the random variable $X_\tau \vert X_0 = x$, is available in closed form as a Gaussian, so obtaining diffused data at any time $\tau$ requires only sampling from this known distribution rather than numerically solving the \gls{sde} \eqref{eq:diffusion}.
 
One of the results by \citet{bounoua2024s} (see also \citet{franzese2023minde, KongBS23}) states that if there is another probability density $q_X$, serving as a reference distribution, for which $q_{X_\tau}$ is generated in the same manner as previously described in \Eqref{eq:diffusion}, then the \gls{kl} divergence between $p_X$ and $q_X$ can be expressed as
\begin{equation}
    \label{eq:kld_diff}
    \KL{p_{X}}{q_{X}} = \int_0^{T} \frac{g_\tau^2}{2} \E{x}{p_{X_\tau}} \left[\left \lVert \Score{p}{X_\tau}(x) - \Score{q}{X_\tau}(x) \right \rVert^2 \right] d\tau + \KL{p_{X_T}}{q_{X_T}},
\end{equation}
where $\lVert \cdot \rVert$ denotes the standard Euclidean norm. This result provides a way to link the \gls{kl} divergence with diffusion processes, given the knowledge of the score functions of $p_{X_\tau}$ and $q_{X_\tau}$. Nonetheless, the availability of such objects is out of reach in practice, and that is why this work instead considers parametric approximations of scores. Thus, for a generic distribution $p$, its score $\Score{p}{X}(x)$ is approximated by a neural network $\Score{p}{X}(x; \theta^\star)$ where $\theta^\star$ is obtained by minimizing the denoising score matching loss \citep{vincent2011connection, song2020score}.
 
Following the work of \citet{franzese2023minde}, we adopt the quantity $e(p, q)$ as an estimator of the \gls{kl} divergence between $p$ and $q$, with
\begin{equation}
\label{eq:kld_estimator}
e(p, q) = \int_0^{T} \frac{g_\tau^2}{2}  \E{x}{p_{X_\tau}} \left[\left \lVert \Score{p}{X_\tau}(x; \theta_1^\star) - \Score{q}{X_\tau}(x; \theta_2^\star) \right \rVert^2 \right] d\tau.
\end{equation}
This is simply the first term of \Eqref{eq:kld_diff}, where parametric scores are used instead of the true score functions. Under the assumption that the learned scores are sufficiently accurate, the terminal \gls{kl} divergence $\KL{p_{X_T}}{q_{X_T}}$ becomes negligible for large $T$, and thus $e(p, q) \approx \KL{p}{q}$.
\subsection{Score-based MI estimation}
\label{sec:score_mi}
Recall from \Secref{sec:MI} that \gls{mi} between two random variables $X$ and $Y$ is defined as $I(X; Y) = \KL{p_{X,Y}}{p_X \, p_Y}$. The main result in \citet{franzese2023minde} provides an accurate way to estimate the \gls{kl} divergence between two densities $p$ and $q$ using a variety of representations of the \gls{mi}, in particular in this work we leverage the representation 
\begin{equation}
\label{eq:mi_rep}
    I(X; Y) = \E{x}{p_X}  \KL{p_{Y_x}}{p_Y},
\end{equation}
where $Y_x$ denotes the random variable $Y \vert X = x$. This choice of representation grants non-negative estimation of \gls{mi} and minimizes the amount of scores to be learned.  

As illustrated in \Eqref{eq:mi_rep}, \gls{mi} can be represented as an expected \gls{kl} divergence, so it can be estimated through \Eqref{eq:kld_estimator}. This task requires learning the score of the conditional density $p_{Y_x}$ and the score of the marginal distribution $p_Y$. A key observation that has been employed in \citep{franzese2023minde, bounoua2024s}, is that all the required scores can be approximated by a single amortized network. The network receives the concatenated input $[X, Y]$ together with a mask vector that encodes the role of each block: blocks whose score is being learned are diffused, blocks serving as conditioning signals retain their clean values, and the remaining blocks are marginalized out. 
\subsection{Estimating the nine MI terms for $\Phi$ID}
\label{sec:masking}
The computation of \gls{phiid} via \gls{mmi} redundancy requires nine distinct \gls{mi} terms of the form $I(\mathbf{a}; \mathbf{b})$, where $\mathbf{a}$ is a non-empty subset of the present-time variables $\{X_{1,t}, X_{2,t}\}$ and $\mathbf{b}$ is a non-empty subset of the future-time variables $\{X_{1,t+1}, X_{2,t+1}\}$. Following the representation in \Eqref{eq:mi_rep}, each such term requires learning the score of a conditional and a marginal density. A single score network is trained applying the masking scheme described in \Secref{sec:score_mi} to the four-block input $[X_{1,t}, \ X_{2,t}, \ X_{1,t+1}, \ X_{2,t+1}]$, so that one model simultaneously approximates all score functions needed for the computation of the nine \gls{mi}s required for the \gls{phiid} atoms computation. Additional implementation details are provided in \Secref{app:implementation}.
\subsection{From nine MI estimates to sixteen $\Phi$ID atoms}
\label{sec:mi_to_atoms}
Once the nine \gls{mi} estimates $\hat{I}(\mathbf{a}; \mathbf{b})$ are obtained, the sixteen \gls{phiid} atoms are recovered through the two-step procedure described in \Secref{sec:phiid}. First, the double redundancy function is evaluated on the product lattice $\mathcal{A}_\Phi$ using \Eqref{eq:double_redundancy}; for the \gls{mmi} redundancy this amounts to selecting the minimum estimated \gls{mi} over the candidate sub-collections for each lattice element:
\begin{equation}
    \label{eq:double_redundancy_hat}
    \hat{I}_\cap^{\Phi}(\alpha, \beta) = \min_{\mathbf{a} \in \alpha} \min_{\mathbf{b} \in \beta} \hat{I}(\mathbf{a}; \mathbf{b}).
\end{equation}
For instance, $\hat{I}_\cap^{\Phi}(\text{Red}, \text{Red}) = \underset{i, j \in \{1, 2\}}{\min}\left\{\hat{I}(X_{i,t}; X_{j,t+1})\right\}$, while for elements with a singleton source or target collection the redundancy function simply returns the corresponding estimated \gls{mi}.

Second, the atoms are obtained through M\"obius inversion on $\mathcal{A}_\Phi$. Since the M\"obius function of the product lattice factorizes as $\mu_\Phi\left((\alpha, \beta), (\alpha', \beta')\right) = \mu_s(\alpha, \alpha') \cdot \mu_t(\beta, \beta')$ \citep{stanley2012enumerative}, the inversion reduces to processing the sixteen nodes in topological order (bottom to top) and computing each atom as the redundancy value at that node minus the sum of all atoms strictly below it:
\begin{equation}
    \label{eq:atom_inversion}
    \hat{\pi}(\alpha, \beta) = \hat{I}_\cap^{\Phi}(\alpha, \beta) - \sum_{\substack{(\alpha', \beta') \prec (\alpha, \beta)}} \hat{\pi}(\alpha', \beta').
\end{equation}
The entire pipeline thus consists of training a single score network, estimating nine \gls{mi} terms, computing sixteen redundancy values via \Eqref{eq:double_redundancy_hat}, and performing M\"obius inversion using \Eqref{eq:atom_inversion}. It is worth noticing that the estimated atoms satisfy $\sum_{(\alpha, \beta)} \hat{\pi}(\alpha, \beta) = \hat{I}(X_t; X_{t+1})$ by construction, which serves as an internal consistency check.
\subsection{Error propagation through M\"obius inversion}
\label{sec:error_propagation}
Let $\mathbf{m} = (m_0, \ldots, m_8)^\top \in \mathbb{R}^9$ denote the true \gls{mi} vector and $\hat{\mathbf{m}} = \mathbf{m} + \boldsymbol{\varepsilon}$ the estimated vector. The pipeline maps $\mathbf{m}$ to the redundancy values $\mathbf{r} \in \mathbb{R}^{16}$ via \Eqref{eq:double_redundancy_hat}, then to atoms $\boldsymbol{\pi} \in \mathbb{R}^{16}$ via \Eqref{eq:atom_inversion}.
 
\begin{assumption}
\label{ass:generic}
All argmin selections in the \gls{mmi} redundancy computation \eqref{eq:double_redundancy_hat} are strict, i.e., no two arguments to any $\min$ operation are equal.
\end{assumption}
 
Under Assumption~\ref{ass:generic}, each $\min$ selects a unique winner and the map $\mathbf{m} \mapsto \mathbf{r}$ is locally affine. We define the \emph{selection matrix} $D \in \{0, 1\}^{16 \times 9}$ with $D_{k,j} = \partial r_k / \partial m_j$; $k \in \mathcal{A}_\Phi$ and $0 \leq j \leq 8$. Clearly each row has exactly one nonzero entry.
 
\begin{theorem}[Jacobian factorization]
\label{thm:jacobian}
Under Assumption~\ref{ass:generic}, the Jacobian of the map $\mathbf{m} \mapsto \boldsymbol{\pi}$ is $J = (M_s \otimes M_t) \, D \in \mathbb{R}^{16 \times 9}$, where $M_s$ and $M_t$ are the M\"obius inversion matrices of the redundancy lattice for the source and target variables, respectively. If the strict ordering on each of $\{m_0, m_1, m_3, m_4\}$, $\{m_2, m_5\}$, and $\{m_6, m_7\}$ is preserved by the estimated counterparts, then the atom error satisfies $\hat{\boldsymbol{\pi}} - \boldsymbol{\pi} = J \, \boldsymbol{\varepsilon}$ exactly.
\end{theorem}


\begin{theorem}[Integer-valuedness]
\label{thm:integer}
Under Assumption~\ref{ass:generic}, all entries of $J$ lie in $\{0, \pm 1\}$.
\end{theorem}
 
Theorem~\ref{thm:integer} enables a clean characterization of error amplification. We define the \emph{structural amplification factor} of atom $(\alpha, \beta)$ as $\mathcal{E}_{(\alpha,\beta)} := \sum_{j} J_{(\alpha,\beta),j}^{\,2}$, which counts the number of \gls{mi} errors contributing to that atom.
 
\begin{theorem}[Amplification bounds]
\label{thm:bounds}
Under Assumption~\ref{ass:generic}: (i) $\mathcal{E}_{\emph{Red} \to \emph{Red}} = 1$; (ii) $\mathcal{E}_{\emph{Syn} \to \emph{Syn}} \in \{4, 6\}$; (iii) $\mathcal{E}_{(\alpha,\beta)} \leq \mathcal{E}_{\emph{Syn} \to \emph{Syn}}$ for all $(\alpha, \beta) \in \mathcal{A}_\Phi$.
\end{theorem}
 
Thus $\text{Syn} \to \text{Syn}$ is provably the hardest atom to estimate, while $\text{Red} \to \text{Red}$ is always the most stable. Proofs are deferred to \Secref{app:proofs}. 
\section{Synthetic Benchmark}
\label{sec:experiment:synthetic}
We evaluate DIPHINE on Gaussian VAR(1) systems for which the nine \gls{mi} terms and the sixteen \gls{phiid} atoms can be computed analytically via the covariance structure \citep{barrett2015exploration}. This provides exact ground truth against which the estimated quantities can be compared. For all experiments, each reported result corresponds to the average over five seeds, where for every seed a new dataset is generated and the model is reinitialized and retrained from the ground up. Additional experimental details, hyperparameters, ablations on the sample size, and results under MI-preserving transforms that render the data non-Gaussian while leaving the ground-truth atoms invariant are provided in \Secref{app:additional_synthetic}.
 
\subsection{Gaussian VAR(1) validation}
\label{sec:gaussian_validation}
Consider a bivariate VAR(1) process $X_t = A X_{t-1} + \varepsilon_t$ where $X_t = [X_{1,t}, X_{2,t}] \in \mathbb{R}^2$, $A \in \mathbb{R}^{2 \times 2}$ is the autoregressive coefficient matrix, and $\varepsilon_t \sim \mathcal{N}(0, \Sigma_\varepsilon)$ are zero-mean Gaussian innovations, independent across time. The stationary covariance $\Sigma_X$ is obtained by solving the discrete Lyapunov equation $\Sigma_X = A \Sigma_X A^\top + \Sigma_\varepsilon$, and the joint covariance of $(X_t, X_{t+1})$ is given by $\text{Cov}(X_t, X_{t+1}) = \Sigma_X A^\top$. The nine ground-truth \gls{mi} terms are then computed from the appropriate sub-blocks of this $4 \times 4$ joint covariance matrix using the Gaussian \gls{mi} formula.
 
We consider three system configurations that cover qualitatively different coupling regimes: a \textbf{coupled} system in which both components influence each other ($A_{12} \neq 0, A_{21} \neq 0$), a \textbf{one-coupling} system in which the influence is unidirectional ($A_{12} \neq 0, A_{21} = 0$), and a \textbf{decoupled} system in which the components evolve independently ($A_{12} = A_{21} = 0$). The scores are trained with $T = 100{,}000$ time points and the corresponding \gls{mi} terms are computed on a subset of unseen $10{,}000$ observations. DIPHINE achieves \gls{mi} MAE of $0.012$, $0.012$, and $0.01$ for the coupled, one-coupling, and decoupled systems respectively. The \gls{mi} bar charts are reported in \Secref{app:additional_synthetic}. The atom-level accuracy for these systems is also reported in the \Secref{app:additional_synthetic}, where the MAE ranges from $\approx 0.001$ for the Red$\to$Red atom to $\approx 0.018$--$0.036$ for the Syn$\to$Syn atom. It is worth noticing that the Syn$\to$Syn atom consistently exhibits the largest MAE across all three systems, which is predicted by the error propagation analysis in \Secref{sec:error_propagation}. 
\subsection{Scalability to higher dimensions}
\label{sec:scalability}
To assess DIPHINE's ability to handle higher-dimensional systems, we consider bipartite VAR(1) processes in which each component $X_{i,t} \in \mathbb{R}^{d}$ is multidimensional, with $d \in \{3, 5, 10\}$, corresponding to total state dimensions of $6$, $10$, and $20$ respectively. Two system types are considered: a \textbf{sparse coupled} system with weak off-diagonal coupling, and a \textbf{decoupled} system with block-diagonal $A$. Ground truth is computed as in \Secref{sec:gaussian_validation}.
 
\begin{figure}[t]
    \centering
    \includegraphics[height=0.42\textwidth]
    {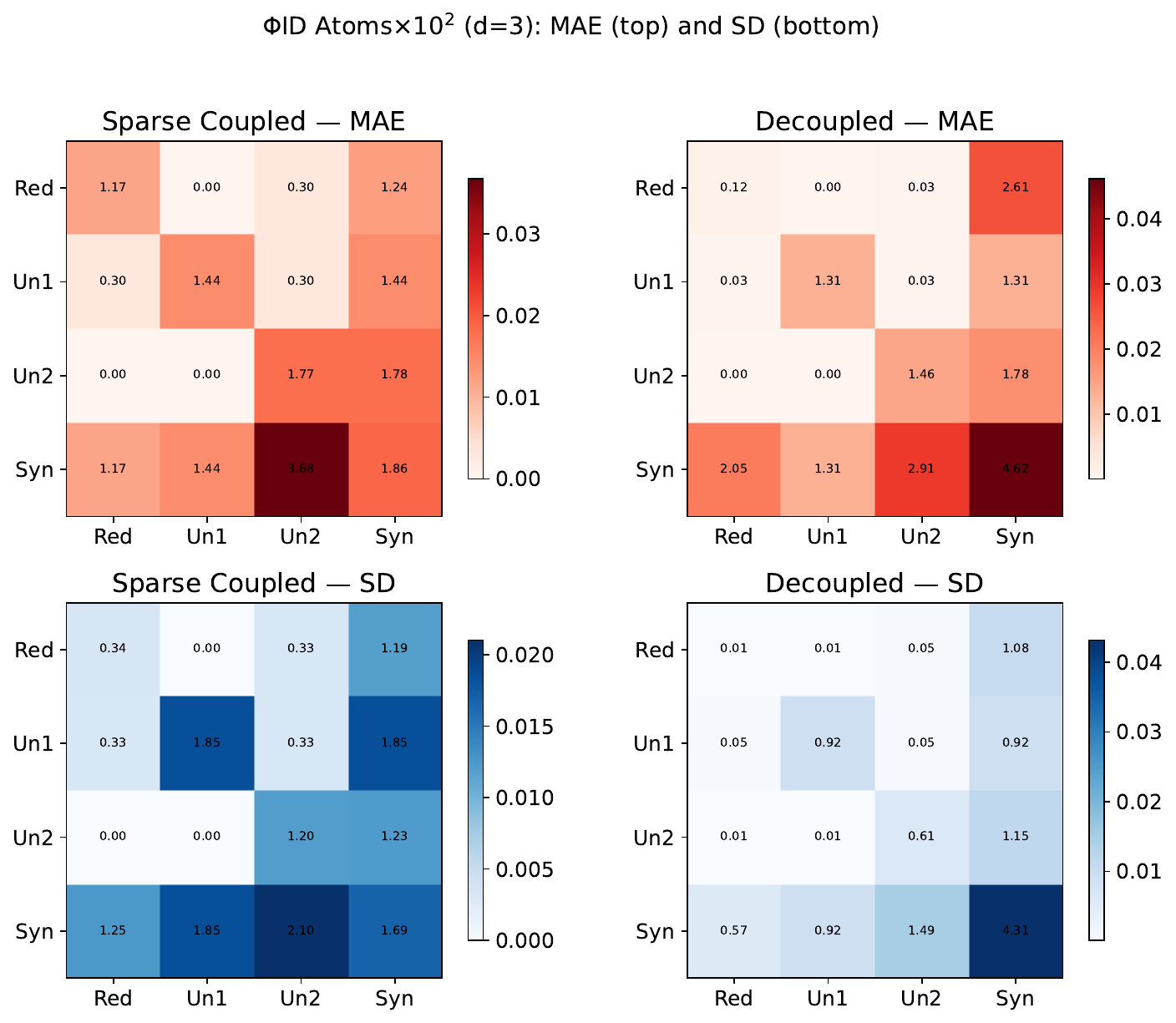}
    \includegraphics[height=0.42\textwidth]
    {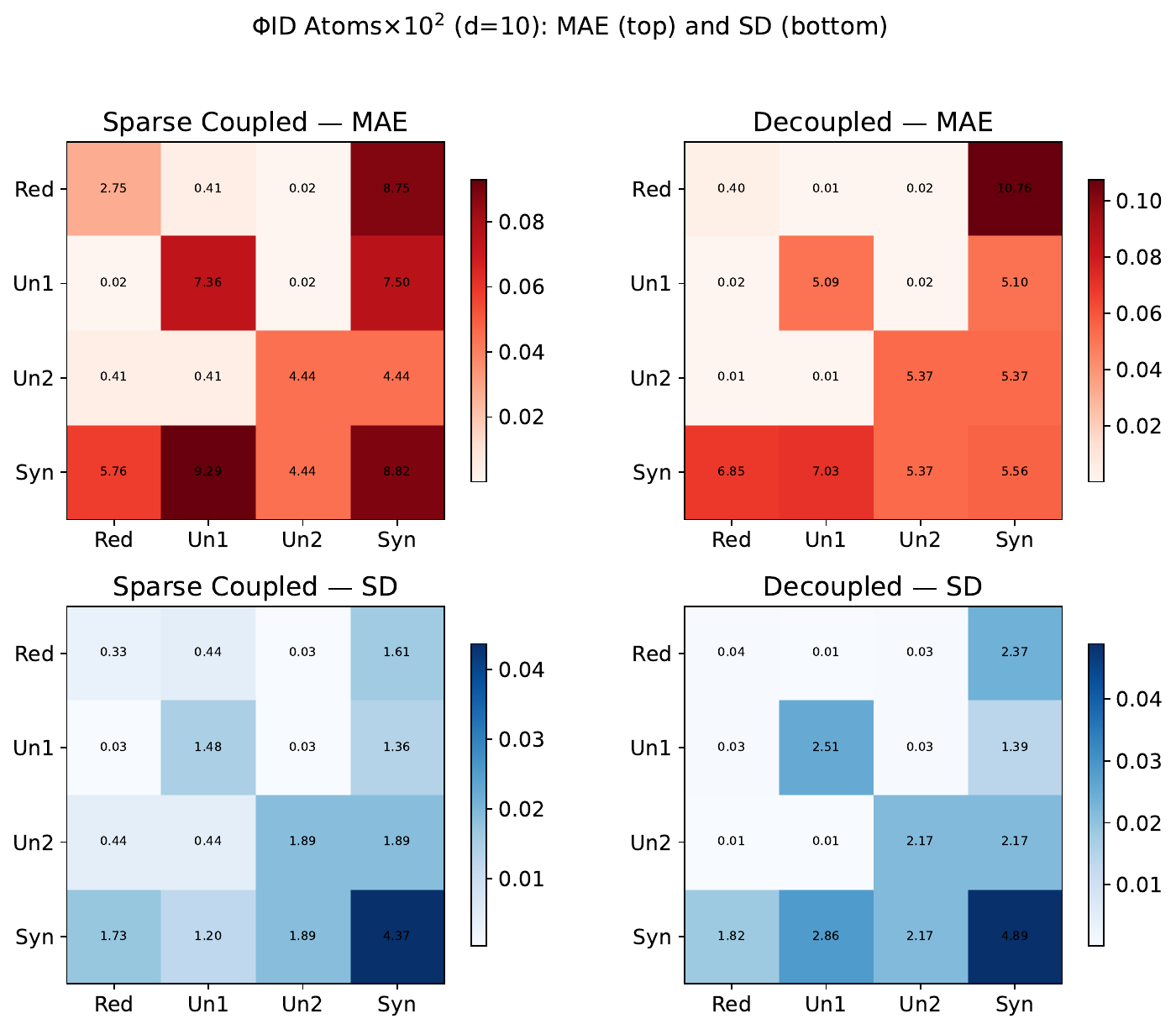}
    \caption{\gls{phiid} atom MAE (top) and standard deviation (bottom) for $d = 3$ (left) and $d = 10$ (right) at $n = 100{,}000$.}
    \label{fig:md_atoms}
\end{figure}

The \gls{mi} MAE increases gradually with dimension: from $0.015$ to $0.018$ at $d = 3$, $0.027$ to $0.033$ at $d = 5$, and $0.058$ to $0.079$ at $d = 10$. \Figref{fig:md_atoms} shows the atom-level MAE for $d = 3$ and $d = 10$. The error pattern is consistent with the $d = 1$ case: atoms involving synergy on both the source and target sides accumulate the largest errors, while atoms at the bottom of the lattice remain well estimated. As a sanity check, the $d = 3$ results match the quality of the $d = 1$ estimates, confirming that the multi-dimensional data pipeline introduces no artifacts. The full set of \gls{mi} bar charts and atom heatmaps for all dimensions is reported in \Secref{app:additional_synthetic}.
\subsection{Comparison with baseline MI estimators}
\label{sec:baselines}
Since no competing method can estimate \gls{phiid} atoms directly, we compare DIPHINE against established \gls{mi} estimators at the level of the nine individual \gls{mi} terms. For each baseline, nine separate estimators are trained (one per \gls{mi} term), and the resulting estimates are passed through the same M\"obius inversion to obtain atoms. The baselines considered are MINE \citep{belghazi2018mine}, NWJ \citep{nguyen2010nwj}, InfoNCE \citep{oord2018infonce}, and KSG \citep{kraskov2004ksg}. This comparison highlights the computational advantage of DIPHINE's single-network approach relative to training nine independent models per baseline.
 
\begin{figure}[t]
    \centering
        \includegraphics[width=0.65\textwidth]{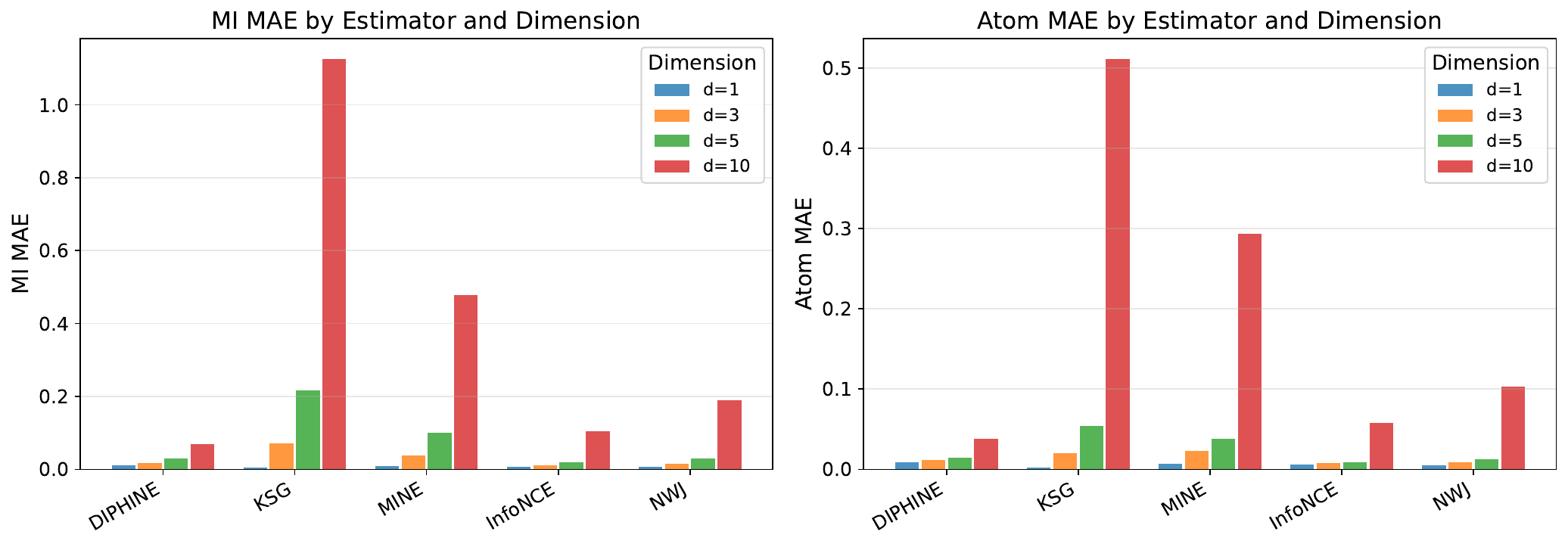}
    \caption{\gls{mi} MAE (left) and atom MAE (right) by estimator and dimension, averaged over all systems and seeds.}
    \label{fig:baselines}
\end{figure}

\begin{figure}[t]
    \centering
        \includegraphics[width=0.65\textwidth]{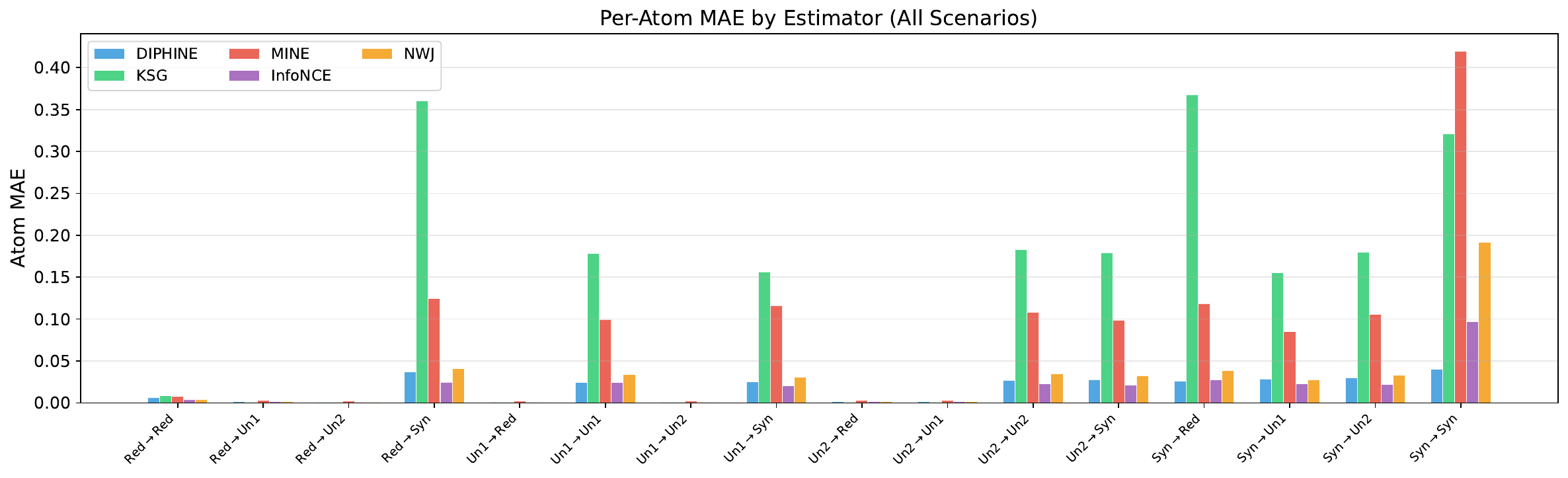}
    \caption{Per-atom MAE by estimator, averaged across all systems and dimensions.}
    \label{fig:per_atom}
\end{figure}

\Figref{fig:baselines} shows the \gls{mi} MAE and atom MAE aggregated across systems and dimensions. DIPHINE achieves the lowest error across all dimensions, with the gap widening as $d$ increases. At $d = 1$ and $d = 3$, NWJ and InfoNCE are competitive at the \gls{mi} level but their atom MAE is consistently higher. KSG degrades sharply beyond $d = 3$, with \gls{mi} MAE approaching $1$ at $d = 10$. MINE exhibits moderate accuracy across dimensions but with a larger dispersion. \Figref{fig:per_atom} provides a finer view, showing the per-atom MAE averaged across all systems and dimensions. DIPHINE achieves the lowest MAE uniformly across all sixteen atoms. The largest discrepancies between methods occur at atoms involving synergy: KSG exhibits atom MAE exceeding $0.35$ at the Red$\to$Syn and Syn$\to$Syn positions, while DIPHINE remains below $0.04$. Atoms at the bottom of the lattice (Red$\to$Red, Red$\to$Un$_1$, Red$\to$Un$_2$) are well estimated by all methods, consistent with the fact that these atoms depend on fewer \gls{mi} terms through the M\"obius inversion.
\section{Real data analysis}
\label{sec:real_data}
We apply DIPHINE to the Fantasia database \citep{iyengar1996age, goldberger2000physiobank}, which consists of continuous recordings from young (21--34 years) and elderly (68--85 years) healthy subjects at rest. For each subject, we extract the RR-interval time series and the respiration signal, forming the bivariate system $X_t = [\text{Resp}_t, \text{RR}_t]$, and compute the full \gls{phiid} decomposition between consecutive time steps. Each result is averaged over 5 seeds.
 
\begin{figure}[t]
    \centering
    \begin{minipage}[t]{0.36\textwidth}
        \centering
        \includegraphics[width=\textwidth]{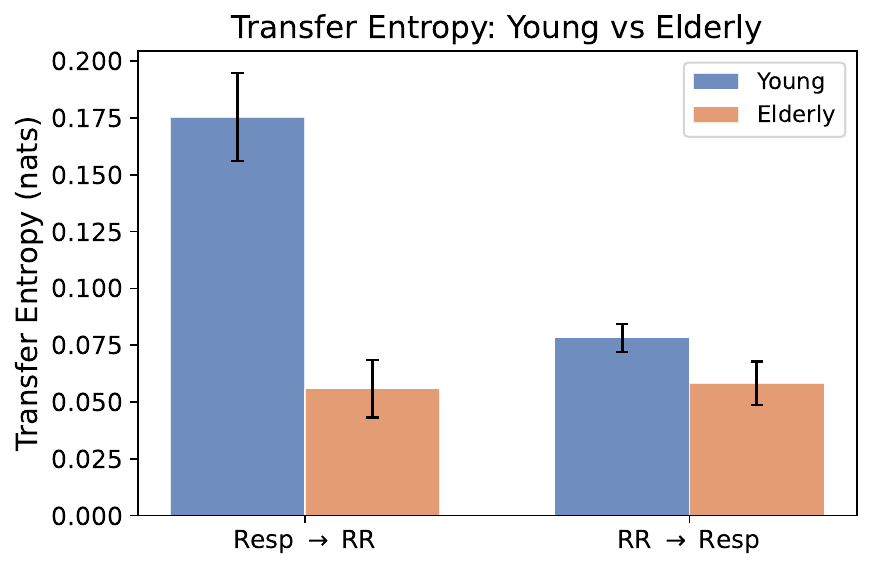}
        \caption{\gls{te} for young and elderly subjects recovered from \gls{phiid} atoms.}
        \label{fig:te_young_elderly}
    \end{minipage}\hfill
    \begin{minipage}[t]{0.62\textwidth}
        \centering
        \includegraphics[width=\textwidth]{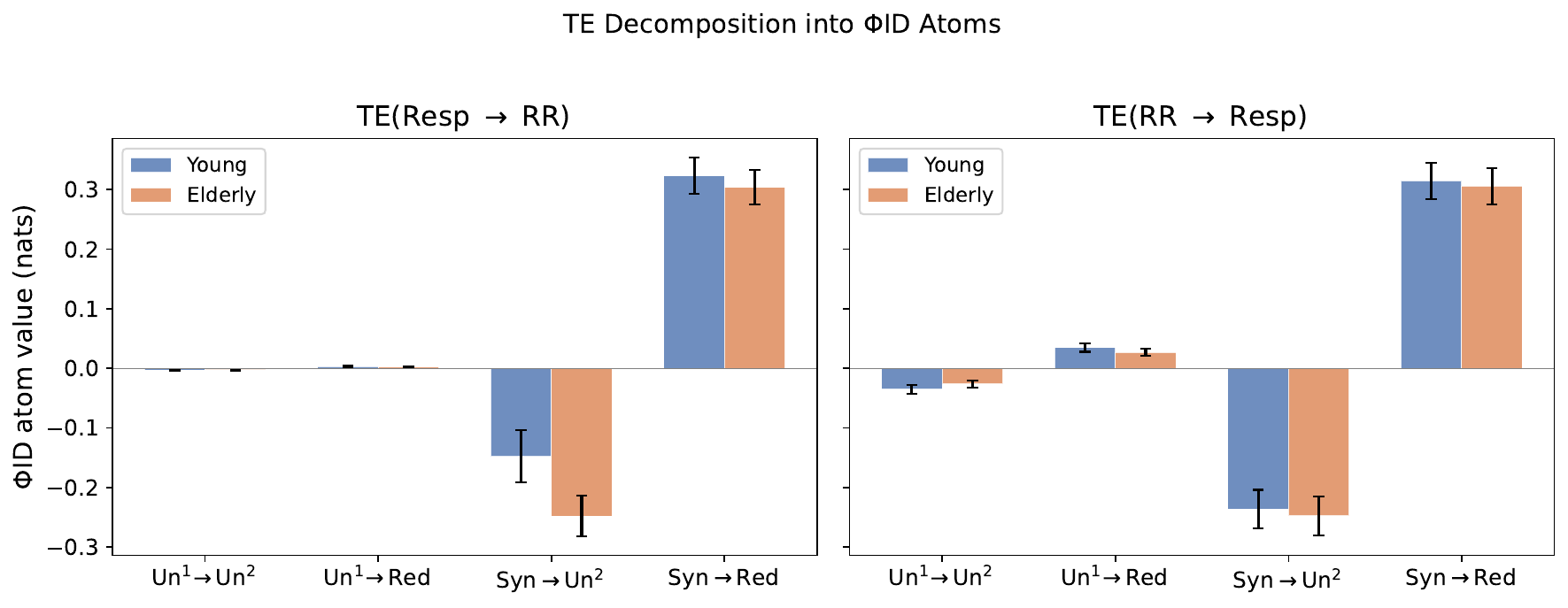}
        \caption{\gls{te} decomposition into \gls{phiid} atoms. Syn$\to$Red dominates in both directions.}
        \label{fig:te_decomposition}
    \end{minipage}
\end{figure}

\paragraph{Transfer entropy and aging.} We recover \gls{te} between respiration and heart rate from the \gls{phiid} atoms. \Figref{fig:te_young_elderly} shows that \gls{te} is higher in young subjects in both directions, with the Resp$\to$RR direction dominant, consistent with the known reduction in parasympathetic cardiorespiratory coupling with aging \citep{iyengar1996age, cactaron2018transfer}. In the elderly, both directions are reduced and the asymmetry is attenuated.
 
\paragraph{\gls{phiid} decomposition.} \Figref{fig:te_decomposition} decomposes the \gls{te} into its constituent atoms. In both directions the Syn$\to$Red atom dominates, indicating that information transfer is synergistic on the source side and redundant on the target side; this pattern is preserved across age groups but reduced in magnitude for the elderly. The full $4 \times 4$ atom matrices, reported in \Secref{app:real_data} (\Figref{fig:app_phiid_heatmaps}), reveal that the largest atoms overall are the self-storage terms Un$_i \to$Un$_i$ and the synergistic integration Syn$\to$Syn. The self-storage atoms increase in the elderly, which may reflect a decoupling of the two subsystems with aging: as cross-variable coupling weakens, each variable's future becomes more predictable from its own past alone. The estimation stability across seeds is also reported in \Secref{app:real_data}.
\section{Conclusions}
\label{sec:conclusions}
Moving beyond the restrictive assumptions of discrete states or Gaussian dynamics is essential for characterizing the rich, fine-grained information dynamics of real-world systems. To address this, we introduced DIPHINE, the first neural estimator of all sixteen \gls{phiid} atoms for continuous systems. Empirically, we demonstrated that a single amortized score network can successfully estimate the nine required \gls{mi} terms, from which the atoms are robustly recovered via M\"obius inversion. Theoretically, we provided a comprehensive error propagation analysis, leveraging the integer-valued nature of the mapping's Jacobian to explicitly bound and characterize the structural amplification of estimation errors across the \gls{phiid} lattice. 

Through extensive evaluation, we showed that DIPHINE matches analytic ground truth on Gaussian VAR(1) systems up to 20 dimensions, outperforms established \gls{mi} baselines, and reveals interpretable information structure in physiological cardio-respiratory data. Current limitations include the restriction to ergodic systems, bivariate systems and the possibility of negative atoms under \gls{mmi} redundancy. Future work includes extension to non-ergodic systems, $N > 2$ partitions (which necessitates navigating a lattice with a super-exponential number of atoms), exploration of alternative redundancy functions, and establishing formal convergence guarantees for the underlying Girsanov-based estimators. Finally, while DIPHINE inherits the computational cost of training the score networks of diffusion models, our amortization strategy across masks effectively mitigates this burden relative to training separate, independent estimators.
\bibliography{refs}
\bibliographystyle{abbrvnat_nourl}


\newpage
\appendix

\section{Construction of the redundancy lattices}
\label{app:lattice_construction}

\subsection{Redundancy lattice}
\label{app:lattice_construction_single}
The \gls{pid} framework formalizes the decomposition in \Eqref{eq:PID} through a lattice structure that organizes all qualitatively distinct ways in which the sources can provide information about the target. To construct such lattice\footnote{Formally, the lattice is the set of antichains of the power set of $\mathcal{S} = \{X_1, X_2\}$, equipped with the partial order defined below.}, one first enumerates the collections of source groups through which information about $Y$ could, in principle, be accessed. In the bivariate case, the possible source groups are $\{X_1\}$, $\{X_2\}$, and $\{X_1, X_2\}$. A source collection is a set of source groups that together specify how the information about $Y$ is accessed. For instance, the collection $\left\{\{X_1\}, \{X_2\}\right\}$ represents information that is available from $X_1$ alone and from $X_2$ alone, that is, redundant information as it could have been obtained from either source. Similarly, the collection $\left\{\{X_1, X_2\}\right\}$ represents information accessible only by observing both sources jointly. However, not all possible collections of source groups should be considered. Each group in a collection asserts that it is sufficient to access the information in question, so if a collection contained both $\{X_1\}$ and $\{X_1, X_2\}$, it would claim that $X_1$ alone suffices and that $\{X_1, X_2\}$ jointly suffices. The latter claim is automatically implied by the former, since any information accessible from $X_1$ alone remains accessible when $X_2$ is additionally observed. Retaining the superset would therefore be logically redundant as it would not specify a genuinely different way of accessing the information. Collections satisfying that no group of sources is a subset of another are called antichains.

For two sources, the resulting antichains are:
\begin{equation}
\label{eq:antichains}
    \underbrace{\{X_1\}\{X_2\}}_{\text{Red}}, \quad
    \underbrace{\{X_1\}}_{\text{Un}_1}, \quad
    \underbrace{\{X_2\}}_{\text{Un}_2}, \quad
    \underbrace{\{X_1, X_2\}}_{\text{Syn}},
\end{equation}
corresponding to redundancy, unique information from $X_1$, unique information from $X_2$, and synergy, respectively. These four antichains are then ordered by the relation
\begin{equation*}
    \alpha \preceq \beta \iff \forall \mathbf{b} \in \beta, \, \exists \, \mathbf{a} \in \alpha, \ \mathbf{a} \subseteq \mathbf{b},
\end{equation*}
which expresses that the information described by $\alpha$ can be accessed through smaller (or equal) groups than that described by $\beta$. Under this ordering, the four antichains form a bounded\footnote{That is, the lattice has a maximal and a minimal elements.} lattice called the redundancy lattice, with $\{X_1\}\{X_2\}$ at the bottom and $\{X_1, X_2\}$ at the top.

A cumulative redundancy function $I_\cap(\alpha; Y)$ is then defined in this lattice to quantify the total information about $Y$ that is accessible through the source pattern described by $\alpha$. The individual \gls{pid} atoms are recovered via M\"obius inversion of $I_\cap$ over the redundancy lattice, as described in \Eqref{eq:mobius_inversion}.

\begin{figure}[h]
    \centering
    \begin{tikzpicture}[
      nd/.style={circle, fill=black, inner sep=2.2pt},
      lbl/.style={rounded corners=3pt, inner sep=3pt, font=\small},
      every path/.style={thick}
    ]
      \node[nd] (red)  at (0,0)    {};
      \node[nd] (un1)  at (-1,1.4) {};
      \node[nd] (un2)  at (1,1.4)  {};
      \node[nd] (syn)  at (0,2.8)  {};
      \node[lbl, fill=bluecol,   below=4pt of red] {\strut Red};
      \node[lbl, fill=orangecol, left=4pt  of un1] {\strut Un$_1$};
      \node[lbl, fill=orangecol, right=4pt of un2] {\strut Un$_2$};
      \node[lbl, fill=redcol,    above=4pt of syn] {\strut Syn};
      \draw (red)--(un1); \draw (red)--(un2);
      \draw (un1)--(syn); \draw (un2)--(syn);
    \end{tikzpicture}
    \caption{The redundancy lattice for two sources. Each node is one of the four antichains in \Eqref{eq:antichains}. Edges connect nodes that are immediately comparable under the partial order $\preceq$: Red is the minimum (most redundant) and Syn is the maximum.}
    \label{fig:redundancy_lattice}
\end{figure}
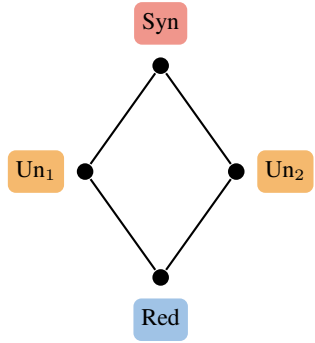

\subsection{Double-redundancy lattice} 
\label{app:lattice_construction_double}
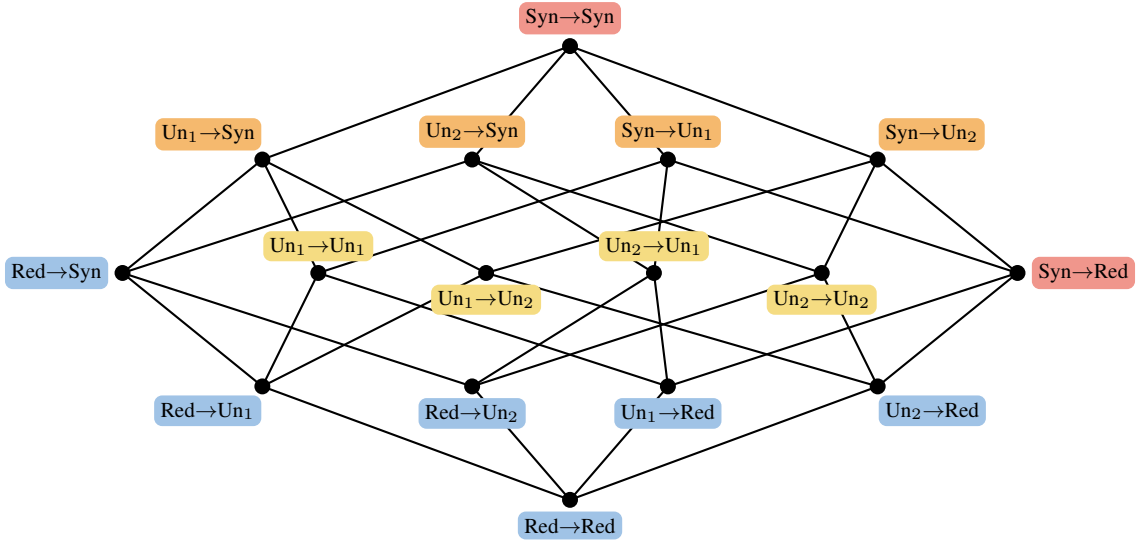
\begin{figure}
    \centering
\begin{tikzpicture}[x=1.85cm, y=1.5cm,
  nd/.style={circle, fill=black, inner sep=2.5pt},
  lbl/.style={rounded corners=3pt, inner sep=2.5pt, font=\footnotesize},
  every path/.style={thick}
]
  \coordinate (rr)   at (0, 0);
  \coordinate (ru1)  at (-2.2, 1);
  \coordinate (ru2)  at (-0.7, 1);
  \coordinate (u1r)  at (0.7, 1);
  \coordinate (u2r)  at (2.2, 1);
  \coordinate (rs)   at (-3.2, 2);
  \coordinate (u1u1) at (-1.8, 2);
  \coordinate (u1u2) at (-0.6, 2);
  \coordinate (u2u1) at (0.6, 2);
  \coordinate (u2u2) at (1.8, 2);
  \coordinate (sr)   at (3.2, 2);
  \coordinate (u1s)  at (-2.2, 3);
  \coordinate (u2s)  at (-0.7, 3);
  \coordinate (su1)  at (0.7, 3);
  \coordinate (su2)  at (2.2, 3);
  \coordinate (ss)   at (0, 4);
  \draw (rr)--(ru1); \draw (rr)--(ru2); \draw (rr)--(u1r); \draw (rr)--(u2r);
  \draw (ru1)--(rs);   \draw (ru1)--(u1u1); \draw (ru1)--(u1u2);
  \draw (ru2)--(rs);   \draw (ru2)--(u2u1); \draw (ru2)--(u2u2);
  \draw (u1r)--(u1u1); \draw (u1r)--(u2u1); \draw (u1r)--(sr);
  \draw (u2r)--(u1u2); \draw (u2r)--(u2u2); \draw (u2r)--(sr);
  \draw (rs)--(u1s);   \draw (rs)--(u2s);
  \draw (u1u1)--(u1s); \draw (u1u1)--(su1);
  \draw (u1u2)--(u1s); \draw (u1u2)--(su2);
  \draw (u2u1)--(u2s); \draw (u2u1)--(su1);
  \draw (u2u2)--(u2s); \draw (u2u2)--(su2);
  \draw (sr)--(su1);   \draw (sr)--(su2);
  \draw (u1s)--(ss);   \draw (u2s)--(ss);
  \draw (su1)--(ss);   \draw (su2)--(ss);
  \foreach \p in {rr,ru1,ru2,u1r,u2r,rs,u1u1,u1u2,u2u1,u2u2,sr,u1s,u2s,su1,su2,ss}
    \fill (\p) circle (3pt);
  \node[lbl, fill=bluecol,   below=4pt]              at (rr)   {Red$\to$Red};
  \node[lbl, fill=bluecol,   below left=3pt and 0pt] at (ru1)  {Red$\to$Un$_1$};
  \node[lbl, fill=bluecol,   below=4pt]              at (ru2)  {Red$\to$Un$_2$};
  \node[lbl, fill=bluecol,   below=4pt]              at (u1r)  {Un$_1$$\to$Red};
  \node[lbl, fill=bluecol,   below right=3pt and 0pt] at (u2r) {Un$_2$$\to$Red};
  \node[lbl, fill=bluecol,   left=5pt]               at (rs)   {Red$\to$Syn};
  \node[lbl, fill=yellcol,   above=4pt]              at (u1u1) {Un$_1$$\to$Un$_1$};
  \node[lbl, fill=yellcol,   below=4pt]              at (u1u2) {Un$_1$$\to$Un$_2$};
  \node[lbl, fill=yellcol,   above=4pt]              at (u2u1) {Un$_2$$\to$Un$_1$};
  \node[lbl, fill=yellcol,   below=4pt]              at (u2u2) {Un$_2$$\to$Un$_2$};
  \node[lbl, fill=redcol,    right=5pt]              at (sr)   {Syn$\to$Red};
  \node[lbl, fill=orangecol, above left=3pt and 0pt] at (u1s)  {Un$_1$$\to$Syn};
  \node[lbl, fill=orangecol, above=4pt]              at (u2s)  {Un$_2$$\to$Syn};
  \node[lbl, fill=orangecol, above=4pt]              at (su1)  {Syn$\to$Un$_1$};
  \node[lbl, fill=orangecol, above right=3pt and 0pt] at (su2) {Syn$\to$Un$_2$};
  \node[lbl, fill=redcol,    above=4pt]              at (ss)   {Syn$\to$Syn};
\end{tikzpicture}
\caption{The double-redundancy lattice for two source and two target variables. Each node is one of the the possible cartesian products in \Eqref{eq:antichains} from the present state to the future state of the system.}
\label{fig:product_lattice}
\end{figure}

The \gls{phiid} framework extends the redundancy lattice to a double-redundancy lattice by taking the Cartesian product of a source lattice $\mathcal{A}_s$ and a target lattice $\mathcal{A}_t$, both isomorphic to the four-element redundancy lattice of \Figref{fig:redundancy_lattice}. The product lattice $\mathcal{A}_s \times \mathcal{A}_t$ contains $4 \times 4 = 16$ elements, ordered componentwise: $(\alpha, \beta) \preceq (\alpha', \beta')$ if and only if $\alpha \preceq \alpha'$ and $\beta \preceq \beta'$. Each element $(\alpha, \beta)$ corresponds to a distinct \gls{phiid} atom, labeled by a pair of information types; one for the source side and one for the target side. \Figref{fig:product_lattice} shows the resulting Hasse diagram. The minimum element Red$\to$Red sits at the bottom and the maximum Syn$\to$Syn at the top, with all 14 mixed atoms arranged at intermediate levels. The double redundancy function $I_\cap^\Phi$ is defined on this lattice via \Eqref{eq:double_redundancy}, and the 16 atoms are recovered by M\"obius inversion on $\mathcal{A}_s \times \mathcal{A}_t$.

\newpage
\section{Proofs}
\label{app:proofs}
\subsection{Proof of Theorem~\ref{thm:jacobian} (Jacobian factorization)}
\label{app:proof_jacobian}
Under Assumption~\ref{ass:generic}, each $\min$ in \Eqref{eq:double_redundancy_hat} admits a unique argmin $j^\star_{(\alpha,\beta)}$. The redundancy map $R$ is locally affine, with $r_{(\alpha,\beta)} = m_{j^\star_{(\alpha,\beta)}}$ in the region of $\mathbb{R}^9$ where the argmin selection is unique, hence $\partial r_{(\alpha,\beta)}/\partial m_j = \mathds{1}_{\{j = j^\star_{(\alpha,\beta)}\}} = D_{(\alpha,\beta),j}$. Since $\boldsymbol{\pi} = M_{\Phi} \mathbf{r}$ is globally linear with $M_{\Phi} = M_s \otimes M_t$ by \citet{stanley2012enumerative}, the chain rule gives $J = M_{\Phi} D$.
For the exact error identity, suppose that $\hat{\mathbf{m}} = \mathbf{m} + \boldsymbol{\varepsilon}$ lies in the same region as $\mathbf{m}$, i.e., the argmin selections do not change. Since $\mathbf{m} \mapsto \boldsymbol{\pi}$ is linear on this region, $\hat{\boldsymbol{\pi}} - \boldsymbol{\pi} = M_\Phi D \boldsymbol{\varepsilon} = J \boldsymbol{\varepsilon}$. \qed
\subsection{Proof of Theorem~\ref{thm:integer} (Integer-valuedness)}
\label{app:proof_integer}
For each $j \in \{0, \ldots, 8\}$ let $C_j \subseteq \mathcal{A}_\Phi$ denote the candidate set of $m_j$, i.e., the nodes of the lattice where $m_j$ enters the minimum in \Eqref{eq:double_redundancy_hat}. Let $S_j = \{(\gamma, \delta) \in C_j : D_{(\gamma, \delta), j} = 1\}$ be the subset where $m_j$ attains that minimum. Clearly $S_j \subseteq C_j$. The $(\alpha, \beta), j$ entry of $J$ reads
\begin{equation}
\label{eq:jacobian_entry}
    J_{(\alpha, \beta), j} = \sum_{(\gamma, \delta) \in S_j} M_{\alpha, \gamma} \, M_{\beta, \delta},
\end{equation}
where we have used $M_s = M_t = M$ on the product lattice. We show that \eqref{eq:jacobian_entry} lies in $\{0, \pm 1\}$ for every realizable $S_j$.

First, the source and target antichains in \Eqref{eq:double_redundancy_hat} are independent, hence $C_j = P_j \times Q_j$ with $P_j \subseteq \mathcal{A}_s$ and $Q_j \subseteq \mathcal{A}_t$. Enumeration yields a $2 \times 2$ grid for $m_0, m_1, m_3, m_4$, a 2-node chain for $m_2, m_5, m_6, m_7$, and a singleton for $m_8$.

Recall that $U \subseteq \mathcal{A}$ is an upset, with $(\mathcal{A}, \preceq)$ a \gls{poset}, if $x \in U$ and $x \preceq y$ then $y \in U$. We claim that winning sets are upsets. Indeed, if $m_j$ wins at $(\gamma, \delta) \in C_j$ and $(\gamma', \delta') \in C_j$ satisfies $(\gamma, \delta) \preceq (\gamma', \delta')$, then the competing \gls{mi}s at $(\gamma', \delta')$ are a subset of those at $(\gamma, \delta)$; hence $m_j$ also wins at $(\gamma', \delta')$. Therefore $S_j$ is an upset in $C_j$.

Now we proceed with the remaining of the proof. For $m_8$ we have $S_8 = \{(\text{Syn}, \text{Syn})\}$, which implies that $J_{(\alpha,\beta), 8} = M_{\alpha, \text{Syn}} \, M_{\beta, \text{Syn}} \in \{0, 1\}$, since the Syn column of $M$ is $(0,0,0,1)^\top$.

For $m_2, m_5, m_6, m_7$, its upsets are $\emptyset$, $\{m_\star\}$, and $C_j$, where $m_\star$ is the maximal element of $C_j$. Observe that all upsets are Cartesian, i.e., $S_j = P'_j \times Q'_j$. Using this fact in \Eqref{eq:jacobian_entry} yields
\begin{equation}
\label{eq:cartesian_factor}
J_{(\alpha,\beta), j} = \left(\sum_{\gamma \in P'_j} M_{\alpha, \gamma}\right) \left(\sum_{\delta \in Q'_j} M_{\beta, \delta}\right),
\end{equation}
and direct evaluation of the factor sums shows that both factors take values in $\{0, \pm 1\}$.

For $m_0, m_1, m_3, m_4$, $C_j$ is a $2 \times 2$ grid with maximal node $m_\star = (\text{Un}_i, \text{Un}_k)$ for appropriate $i, k \in \{1, 2\}$. Its upsets are $\emptyset$; $\{m_\star\}$, $\{m_\star, (\text{Red}, \text{Un}_k)\}$, $\{m_\star, (\text{Un}_i, \text{Red})\}$, $C_j \setminus \{(\text{Red}, \text{Red})\}$, and $C_j$. The first four and the last are Cartesian, hence handled by \eqref{eq:cartesian_factor}. The remaining case $S_j = C_j \setminus \{(\text{Red}, \text{Red})\}$ satisfies
\begin{equation}
\label{eq:non_cartesian}
J_{(\alpha,\beta), j} = \left(M_{\alpha, \text{Un}_i} + M_{\alpha, \text{Red}}\right)\left(M_{\beta, \text{Un}_k} + M_{\beta, \text{Red}}\right) - M_{\alpha, \text{Red}} \, M_{\beta, \text{Red}}.
\end{equation}
Exhaustive evaluation over $(\alpha, \beta) \in \mathcal{A}_\Phi$ yields entries in $\{0, \pm 1\}$. \qed
\subsection{Proof of Theorem~\ref{thm:bounds} (Amplification bounds)}
\label{app:proof_bounds}
By Theorem~\ref{thm:integer}, $J_{(\alpha,\beta), j}^2 \in \{0, 1\}$, so $\mathcal{E}_{(\alpha,\beta)}$ counts the nonzero entries in row $(\alpha, \beta)$ of $J$.

First, we show that $\mathcal{E}_{\emph{\text{Red}} \to \emph{\text{Red}}} = 1$. The Red row of $M$ is $(1, 0, 0, 0)$, hence $M_{\text{Red}, \gamma} \, M_{\text{Red}, \delta}$ is nonzero only if $\gamma = \delta = \text{Red}$. Therefore $J_{(\text{Red}, \text{Red}), j} = \sum_{(\gamma, \delta) \in S_j} M_{\text{Red}, \gamma} \, M_{\text{Red}, \delta}$ is nonzero only for the unique $j$ such that $m_j$ wins $\min(m_0, m_1, m_3, m_4)$, in which case it equals 1. Hence the Red-Red row of $J$ has a single nonzero entry, so $\mathcal{E}_{\emph{\text{Red}} \to \emph{\text{Red}}} = 1$.

Now let us show that $\mathcal{E}_{\text{\emph{Syn}} \to \text{\emph{Syn}}} \in \{4, 6\}$. The Syn row of $M$ is $(1, -1, -1, 1)$, so every entry of the $(\text{Syn}, \text{Syn})$ row of $M_{\Phi}$ is $\pm 1$ and the contribution of each $m_j$ to $J_{(\text{Syn}, \text{Syn}), j}$ is determined by the shape of its winning set.

Applying \eqref{eq:cartesian_factor} and \eqref{eq:non_cartesian} with $\alpha = \beta = \text{Syn}$, and noting $M_{\text{Syn}, \text{Un}_i} + M_{\text{Syn}, \text{Red}} = 0$, the contributions are: 0 if $S_j = C_j$ (full grid or full 2-node chain), $\pm 1$ if $S_j = \{m\star\}$, 0 if $S_j$ is a 2-element Cartesian upset containing a Red coordinate, and $-1$ if $S_j = C_j \setminus \{(\text{Red}, \text{Red})\}$.

The contributions from $m_2, m_5, m_6, m_7, m_8$ are fixed. Among $\{m_2, m_5\}$, exactly one wins $\min(m_2, m_5)$ and contributes $0$; the other loses and contributes $\pm 1$. The same holds for $\{m_6, m_7\}$. Finally, $m_8$ contributes $M_{\text{Syn}, \text{Syn}}^2 = 1$. These yield 3 nonzero contributions in every configuration.

The contributions from $\{m_0, m_1, m_3, m_4\}$ are determined by the strict total ordering of these four \gls{mi}s, of which there are $4! = 24$. Let $j^\star$ denote the 4-way argmin winner, its winning set is the full grid and it contributes $0$. Among the remaining three, denote by $j'$ the diagonal opposite of $j^\star$ in $C_{j^\star}$. If $j'$ is the smallest of the three non-winners, its winning set is $C_{j'} \setminus \{(\text{Red}, \text{Red})\}$ and the two row/column neighbors have winning set $\{m\star\}$, giving three nonzero contributions. Otherwise, $j'$ has winning set $\{m\star\}$ and the two neighbors have winning sets of the form $\{m\star\} \cup \{(\ast, \text{Red})\}$ or $\{m\star\} \cup \{(\text{Red}, \ast)\}$, yielding one nonzero contribution. The first case occurs for $2$ of the $6$ orderings of the non-winners, the second for the remaining $4$.

Combined with the $2 \times 2 = 4$ independent orderings of $(m_2, m_5)$ and $(m_6, m_7)$ and the $4$ choices of $j^\star$, the $96$ configurations partition into $4 \cdot 2 \cdot 4 = 32$ yielding $\mathcal{E}_{\text{Syn} \to \text{Syn}} = 3 + 3 = 6$ and $4 \cdot 4 \cdot 4 = 64$ yielding $\mathcal{E}_{\text{Syn} \to \text{Syn}} = 3 + 1 = 4$.

Finally, we prove that $\mathcal{E}_{(\alpha,\beta)} \leq \mathcal{E}_{\text{\emph{Syn}} \to \text{\emph{Syn}}}$. The $(\text{Syn}, \text{Syn})$ row of $M_{\Phi}$ is the only row with no zero entry, since every other row of $M$ contains at least one zero. For any other atom, at least one column of $D$ is annihilated before summation in \eqref{eq:jacobian_entry}. Exhaustive evaluation over all $96$ configurations confirms that no atom exceeds $\mathcal{E}_{\text{Syn} \to \text{Syn}}$, with equality possible only when $\mathcal{E}_{\text{Syn} \to \text{Syn}} = 4$. \qed
\newpage
\section{Implementation details}
\label{app:implementation}
 
\subsection{Masking scheme}
\label{app:masks}
As described in \Secref{sec:masking}, DIPHINE operates on the concatenated state $[X_{1,t}, X_{2,t}, X_{1,t+1}, X_{2,t+1}]$ using a four-block mask $m = [m_1, m_2, m_3, m_4]$ with entries in $\{-1, 0, 1\}$: $m_i = 1$ indicates the block is diffused and its score is learned, $m_i = 0$ indicates the block serves as a conditioning signal (clean data), and $m_i = -1$ indicates the block is marginalized out. Each of the nine \gls{mi} terms requires a conditional mask and a marginal mask. The 9 conditional masks are:
\begin{align*}
&[0, -1, 1, -1], \quad [0, -1, -1, 1], \quad [0, -1, 1, 1], \\
&[-1, 0, 1, -1], \quad [-1, 0, -1, 1], \quad [-1, 0, 1, 1], \\
&[0, 0, 1, -1], \quad [0, 0, -1, 1], \quad [0, 0, 1, 1],
\end{align*}
corresponding to $I(X_{1,t}; X_{1,t+1})$, $I(X_{1,t}; X_{2,t+1})$, $I(X_{1,t}; X_{t+1})$, $I(X_{2,t}; X_{1,t+1})$, $I(X_{2,t}; X_{2,t+1})$, $I(X_{2,t}; X_{t+1})$, $I(X_t; X_{1,t+1})$, $I(X_t; X_{2,t+1})$, and $I(X_t; X_{t+1})$ respectively. The marginal masks are derived by replacing all conditioning entries ($0$) with marginalization entries ($-1$), yielding 3 unique marginal masks: $[-1, -1, 1, -1]$, $[-1, -1, -1, 1]$, and $[-1, -1, 1, 1]$. The total number of unique masks is 12.
 
\subsection{Architecture and training}
\label{app:architecture}
The score network is a Multi-layer Perceptron conditioned on the diffusion time $\tau$ and the mask vector $m$, with hidden dimension $128$ for $2d \leq 50$ and $192$ otherwise. We use the \gls{vpsde} with importance sampling. Training uses Adam at learning rate $10^{-3}$, batch size $256$, for $500$ epochs with \gls{ema} decay $0.999$. At inference, the time integral in \Eqref{eq:kld_estimator} is approximated by Monte Carlo sampling over $10$ time points.
\newpage
\section{Experimental setup and additional synthetic results}
\label{app:additional_synthetic}
\subsection{VAR(1) system configurations}
\label{app:var_systems}
For the $d = 1$ experiments, the three system configurations use the following parametrization:
\begin{equation*}
A_{\text{coupled}} = \begin{pmatrix} 0.7 & 0.6 \\ -0.3 & 0.8 \end{pmatrix}, \quad
A_{\text{one-coupling}} = \begin{pmatrix} 0.7 & 0.0 \\ -0.3 & 0.8 \end{pmatrix}, \quad
A_{\text{decoupled}} = \begin{pmatrix} 0.7 & 0.0 \\ 0.0 & 0.8 \end{pmatrix},
\end{equation*}
all with innovation covariance $\Sigma_\varepsilon = \begin{bmatrix} 1 & 0.3 \\ 0.3 & 1 \end{bmatrix}$. The spectral radii are $0.99$, $0.80$, and $0.80$ respectively, ensuring stationarity. The systems are simulated with $T = 100{,}000$ time steps after a burn-in of $2{,}000$ steps, and inference is performed on a held-out set of $10{,}000$ observations.
 
For the multi-dimensional experiments ($d \in \{3, 5, 10\}$), each diagonal block of $A$ is a random rotation scaled to a target spectral radius of $0.85$. The off-diagonal coupling blocks contain sparse entries of magnitude $0.15$ for the sparse coupled configuration and $0$ for the decoupled configuration. Innovation covariance has within-block correlation $0.2$ and no cross-block correlation.
\subsection{MI bar charts ($d = 1$)}
\label{app:d1_mi}
\begin{figure}[H]
    \centering
    \includegraphics[width=\textwidth]{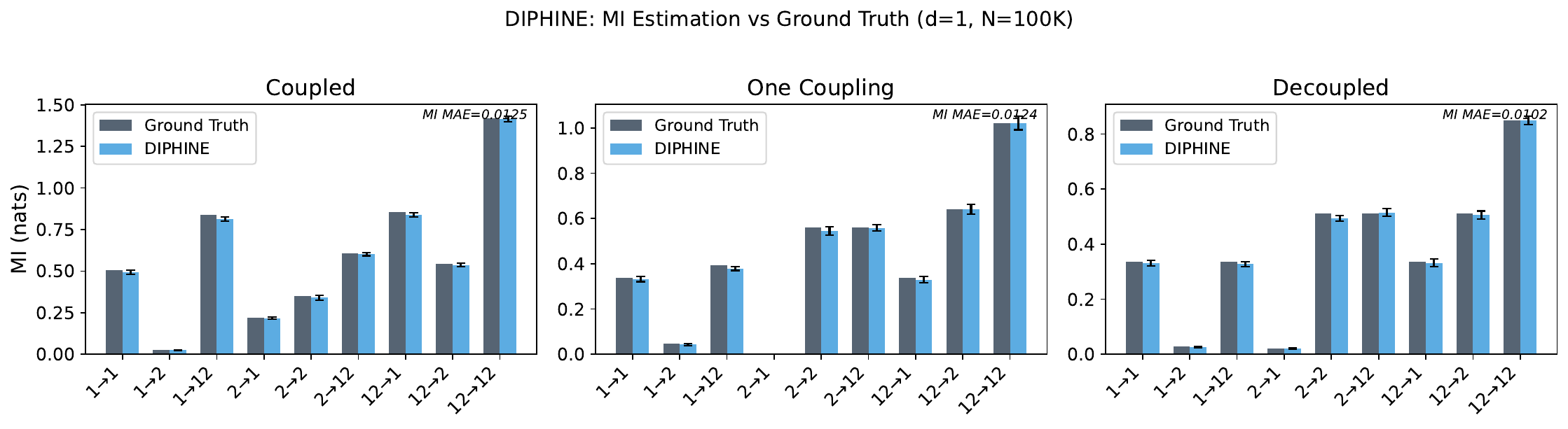}
    \caption{Estimated \gls{mi} values versus analytic ground truth for three Gaussian VAR(1) systems at $d = 1$, $n = 100{,}000$. Error bars correspond to standard deviations over 5 seeds.}
    \label{fig:app_d1_mi}
\end{figure}
\subsection{Atom heatmaps ($d = 1$)}
\label{app:d1_atoms}
\begin{figure}[H]
    \centering
    \includegraphics[width=\textwidth]{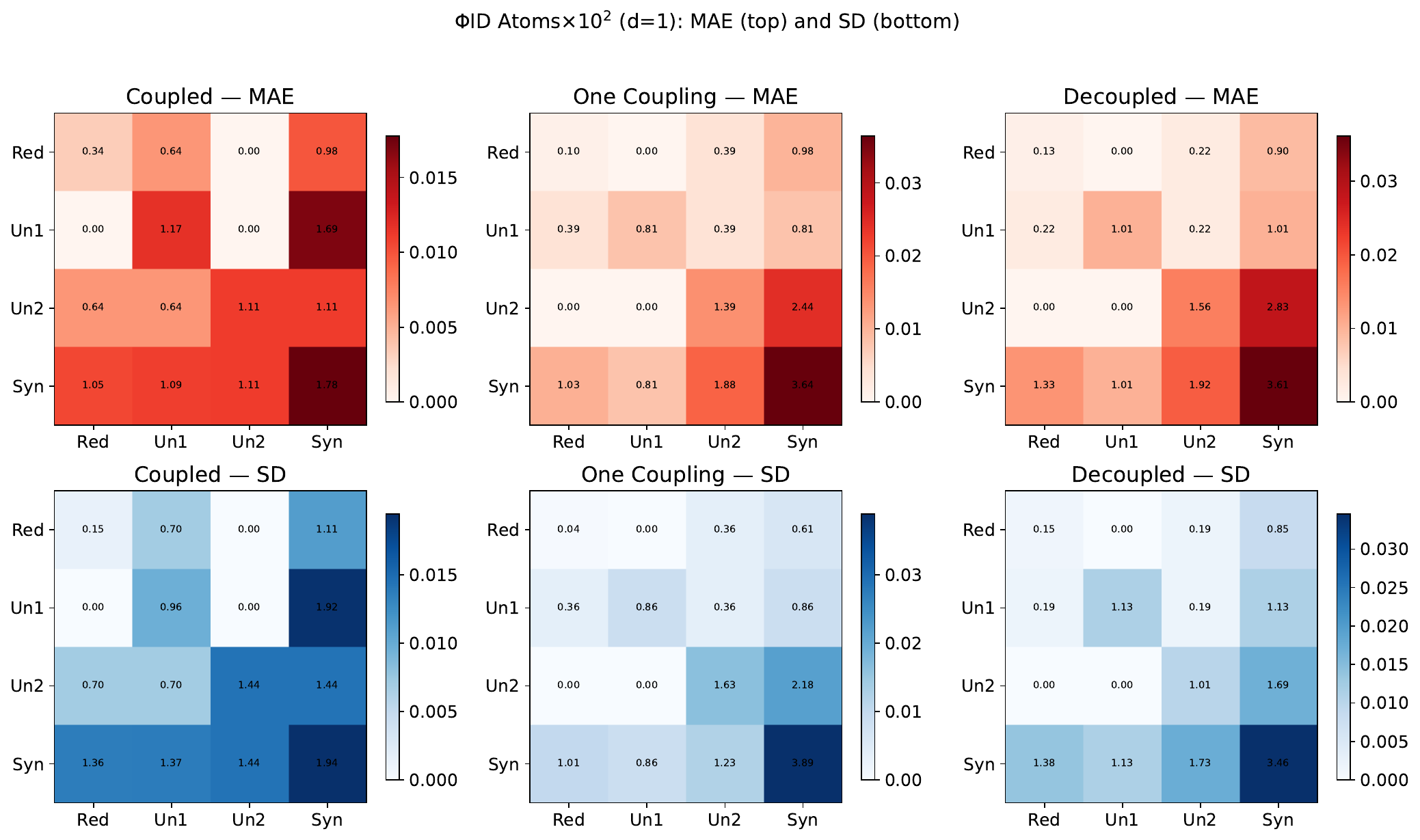}
    \caption{\gls{phiid} atom MAE (top row) and standard deviation (bottom row) for the three $d = 1$ systems at $n = 100{,}000$. The Syn$\to$Syn atom consistently exhibits the largest MAE, as predicted by the error propagation analysis in \Secref{sec:error_propagation}. The Red$\to$Red atom is the most accurately estimated across all systems.}
    \label{fig:app_d1_atoms}
\end{figure}
\subsection{MI bar charts ($d = 3, 5, 10$)}
\label{app:md_mi}
\begin{figure}[H]
    \centering
    \includegraphics[width=\textwidth]{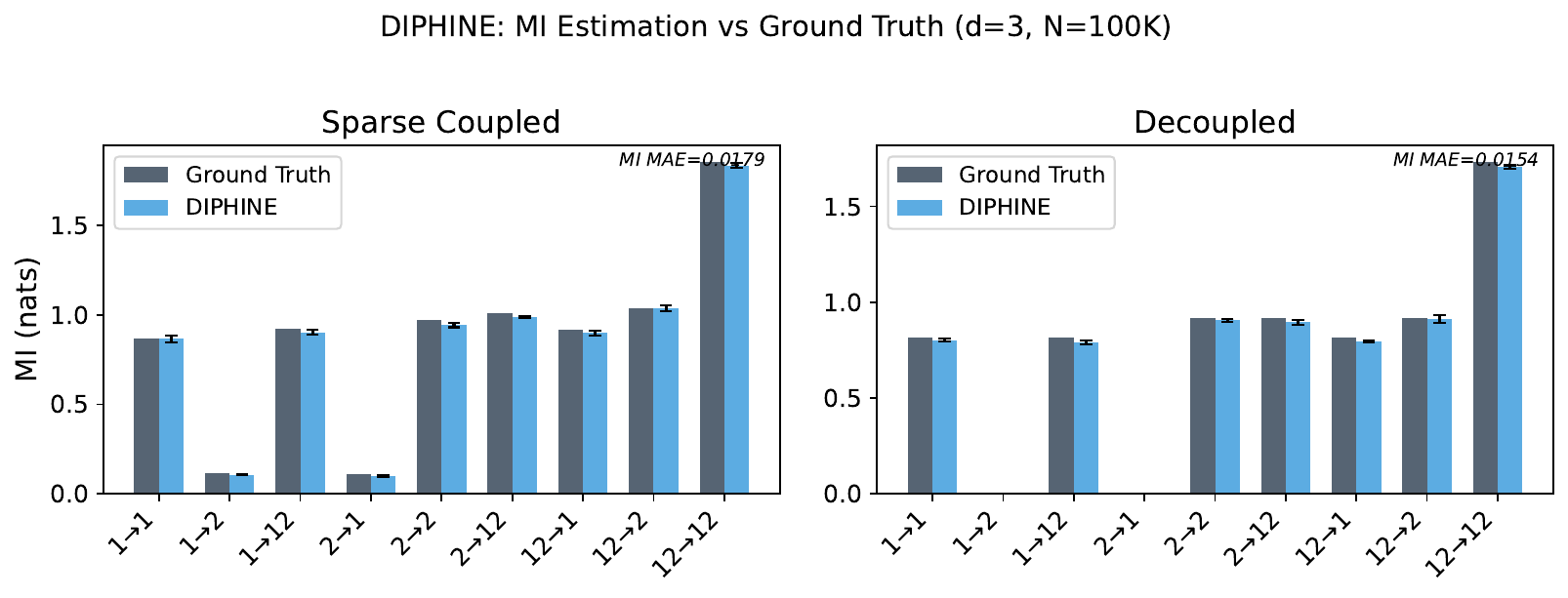}
    \caption{Estimated \gls{mi} values versus ground truth for $d = 3$ (sparse coupled and decoupled) at $n = 100{,}000$.}
    \label{fig:app_d3_mi}
\end{figure}
\begin{figure}[H]
    \centering
    \includegraphics[width=\textwidth]{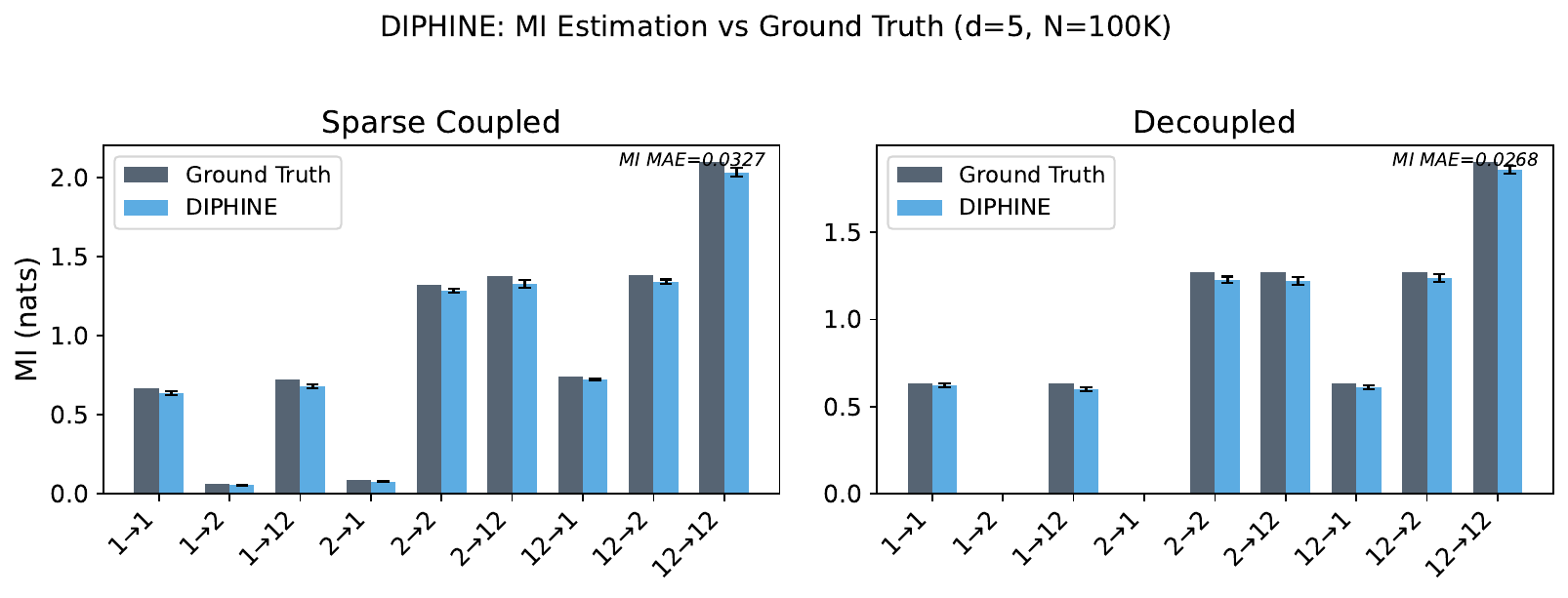}
    \caption{Estimated \gls{mi} values versus ground truth for $d = 5$ at $n = 100{,}000$. The \gls{mi} MAE is $0.033$ and $0.027$ for the sparse coupled and decoupled systems.}
    \label{fig:app_d5_mi}
\end{figure}
\begin{figure}[H]
    \centering
    \includegraphics[width=\textwidth]{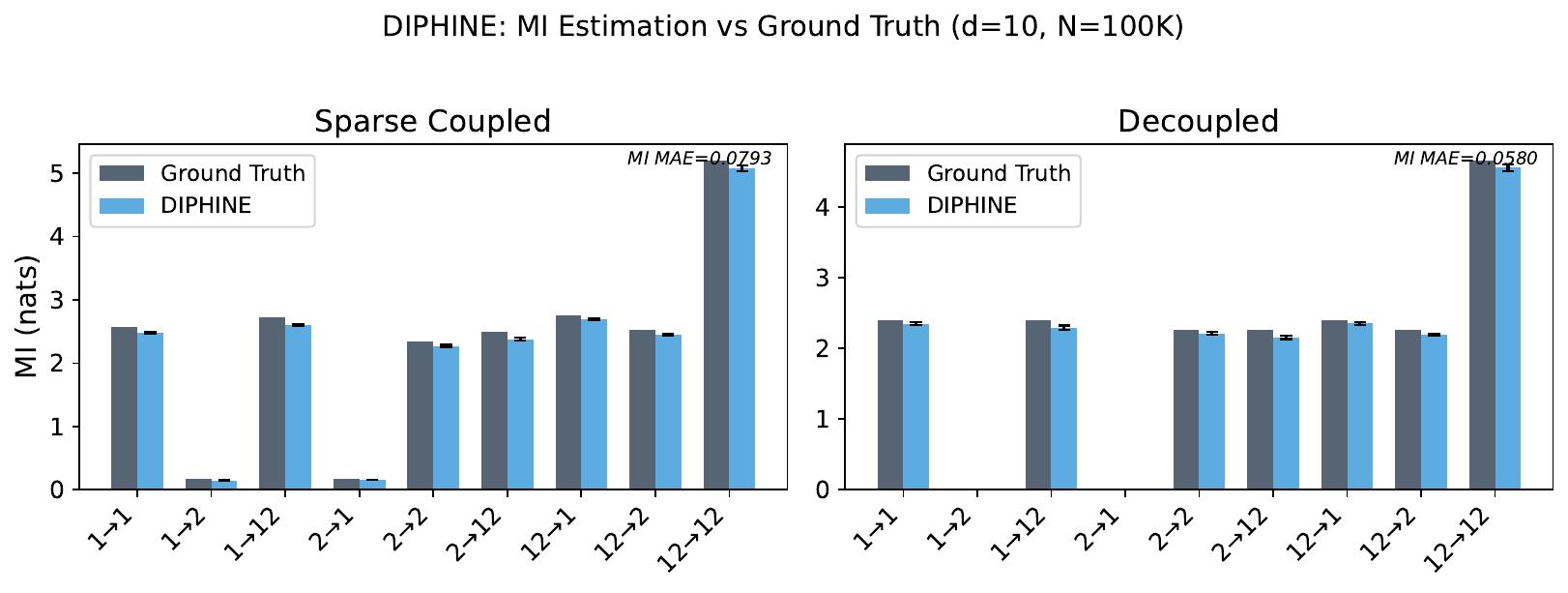}
    \caption{Estimated \gls{mi} values versus ground truth for $d = 10$ at $n = 100{,}000$. The \gls{mi} MAE increases to $0.079$ and $0.058$, reflecting the higher difficulty of score estimation in 20-dimensional systems.}
    \label{fig:app_d10_mi}
\end{figure}
\subsection{Atom heatmaps ($d = 5$)}
\label{app:d5_atoms}
\begin{figure}[H]
    \centering
    \includegraphics[width=\textwidth]{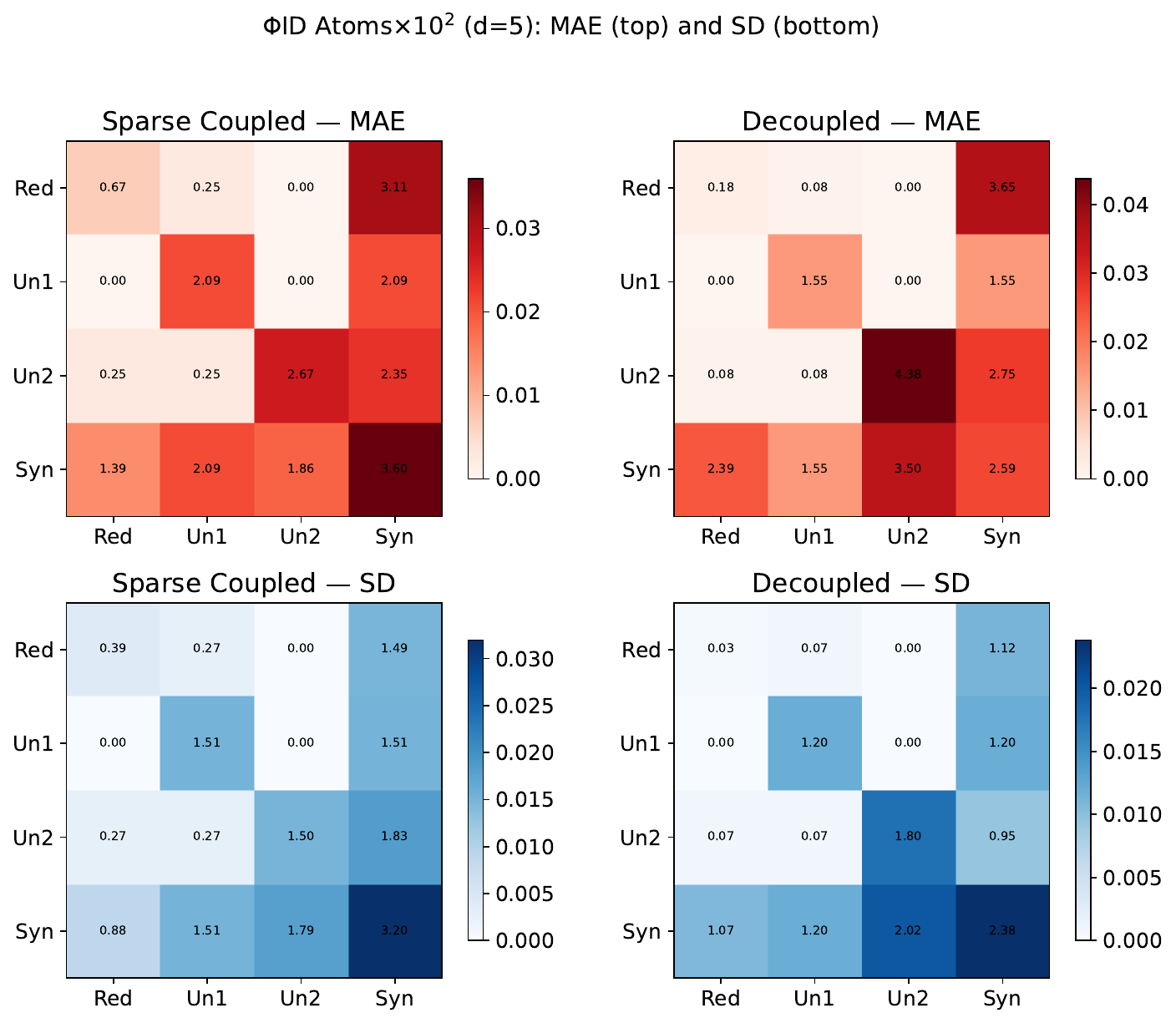}
    \caption{\gls{phiid} atom MAE and standard deviation for $d = 5$ at $n = 100{,}000$. The error pattern is consistent with $d = 1$ and $d = 3$: atoms involving synergy on both sides accumulate the largest errors.}
    \label{fig:app_d5_atoms}
\end{figure}
\newpage
\section{MI-preserving transforms}
\label{app:transforms}
To demonstrate that DIPHINE extends beyond the Gaussian regime while retaining exact ground truth for validation, we apply componentwise invertible transformations to the Gaussian VAR(1) data. Since these transformations act independently on each scalar component, they preserve all \gls{mi} values and therefore all \gls{phiid} atoms, while rendering the marginal distributions non-Gaussian. We consider two transformations: the \textbf{half-cube} transform $h(x) = x |x|^{1/2}$, which expands the tails of the distribution, and the \textbf{CDF} transform $\Phi(x)$ where $\Phi$ is the empirical CDF of the standard Gaussian distribution.
 
\begin{figure}[H]
    \centering
    \includegraphics[width=\textwidth]{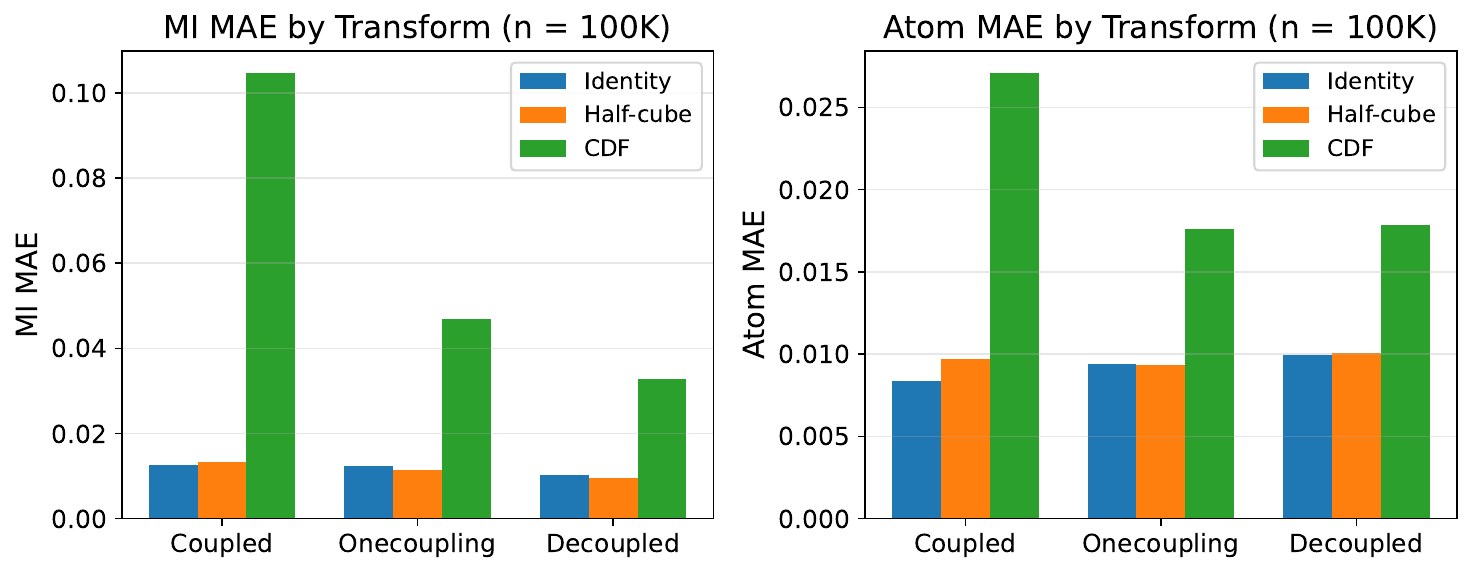}
    \caption{\gls{mi} MAE (left) and atom MAE (right) under identity, half-cube, and CDF transforms for the three $d = 1$ systems at $n = 100{,}000$. The half-cube transform yields accuracy comparable to the identity case, while the CDF transform is more challenging but preserves the qualitative decomposition structure.}
    \label{fig:app_transforms}
\end{figure}

\Figref{fig:app_transforms} shows the \gls{mi} MAE and atom MAE at $n = 100{,}000$ for both transformations compared to the identity (untransformed) case. Under the half-cube transform, the accuracy is comparable to the identity case across all three systems, confirming that the score network adapts to the altered marginal structure without loss of accuracy. The CDF transform is more challenging: the \gls{mi} MAE increases, with the coupled system being the most affected. Nevertheless, the atom MAE remains moderate, and the qualitative structure of the decomposition is preserved. These results demonstrate that DIPHINE can operate in non-Gaussian settings where existing analytic methods are inapplicable.
\newpage
\section{Ablation studies}
\label{app:ablations}
\subsection{Sample size ablation}
\label{app:sample_size}
\begin{figure}[H]
    \centering
    \includegraphics[width=\textwidth]{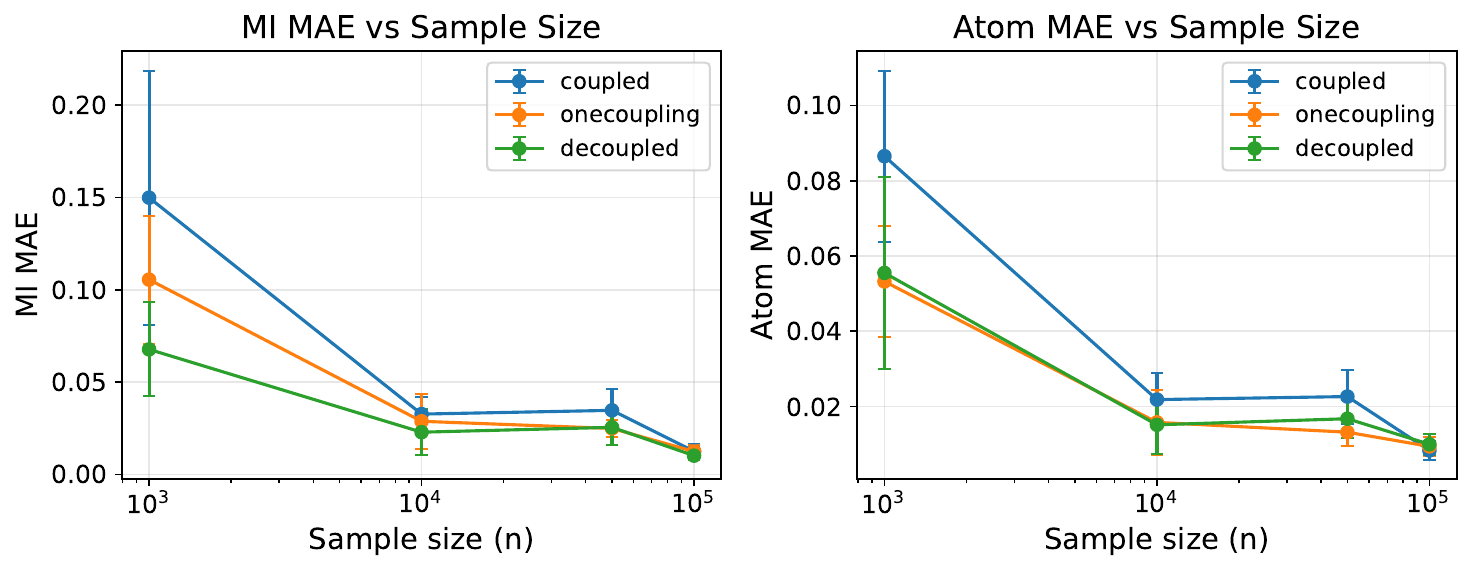}
    \caption{\gls{mi} MAE (left) and atom MAE (right) as a function of sample size for the three $d = 1$ Gaussian VAR(1) systems. Convergence is monotonic, with the largest gains between $n = 1{,}000$ and $n = 10{,}000$.}
    \label{fig:app_sample_size}
\end{figure}

To assess the effect of the training set size on estimation accuracy, we evaluate DIPHINE on the three $d = 1$ systems at $n \in \{1{,}000, \ 10{,}000, \ 50{,}000, \ 100{,}000\}$. \Figref{fig:app_sample_size} shows the \gls{mi} MAE and atom MAE as a function of $n$. Convergence toward the ground truth is clearly visible as the sample size increases, with the largest improvements occurring between $n = 1{,}000$ and $n = 10{,}000$. By $n = 100{,}000$, all three systems achieve \gls{mi} MAE below $0.013$.
\subsection{Coupling strength sweep}
\label{app:coupling_md}
\begin{figure}[H]
    \centering
    \includegraphics[width=\textwidth]{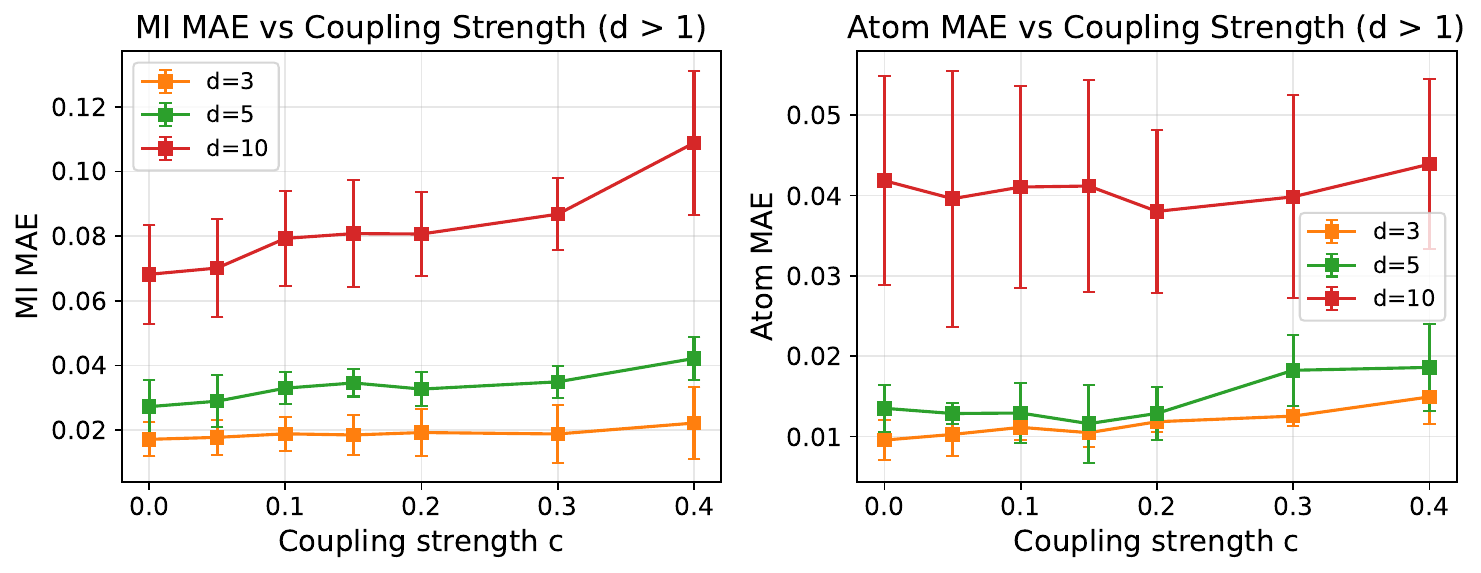}
    \caption{DIPHINE \gls{mi} MAE (left) and atom MAE (right) for $d \in \{3, 5, 10\}$ at $n = 100{,}000$.}
    \label{fig:app_coupling_md}
\end{figure}

\begin{figure}[H]
    \centering
    \includegraphics[width=\textwidth]{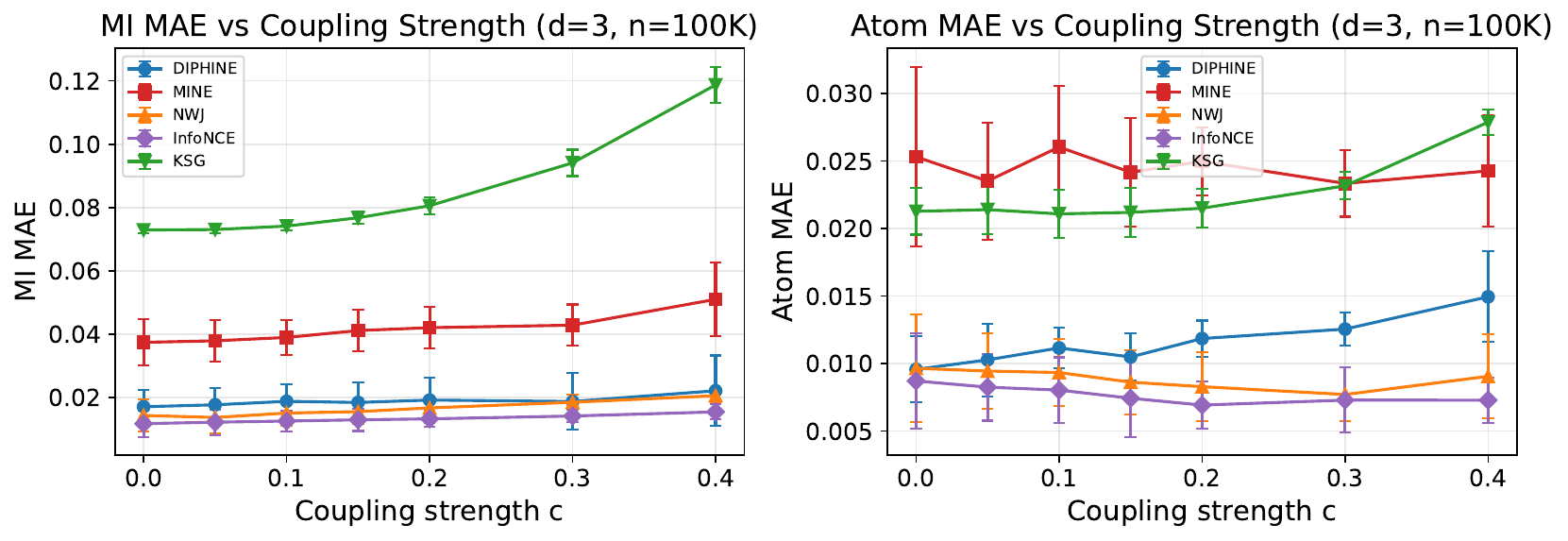}
    \caption{All methods at $d = 3$, $n = 100{,}000$. InfoNCE and NWJ remain competitive with DIPHINE; KSG degrades substantially.}
    \label{fig:app_coupling_d3_all}
\end{figure}

\begin{figure}[H]
    \centering
    \includegraphics[width=\textwidth]{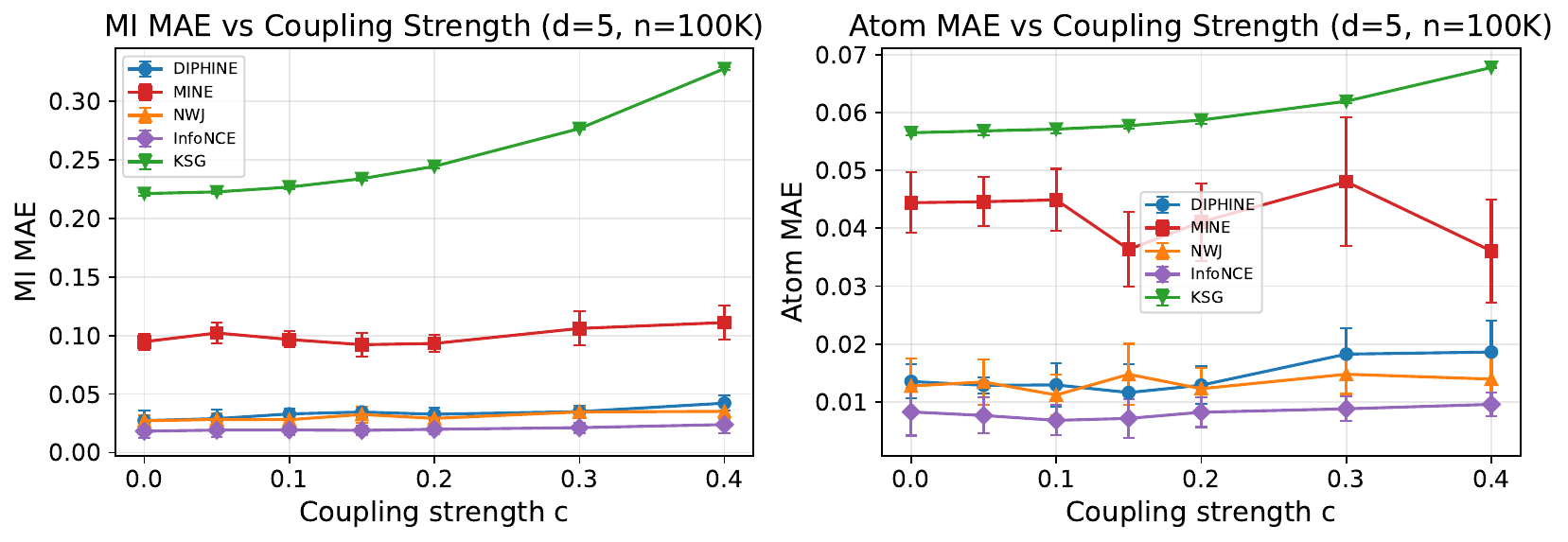}
    \caption{All methods at $d = 5$, $n = 100{,}000$. InfoNCE achieves lower MI MAE than DIPHINE, but DIPHINE pulls ahead on atom MAE. KSG deteriorates further.}
    \label{fig:app_coupling_d5_all}
\end{figure}

\begin{figure}[H]
    \centering
    \includegraphics[width=\textwidth]{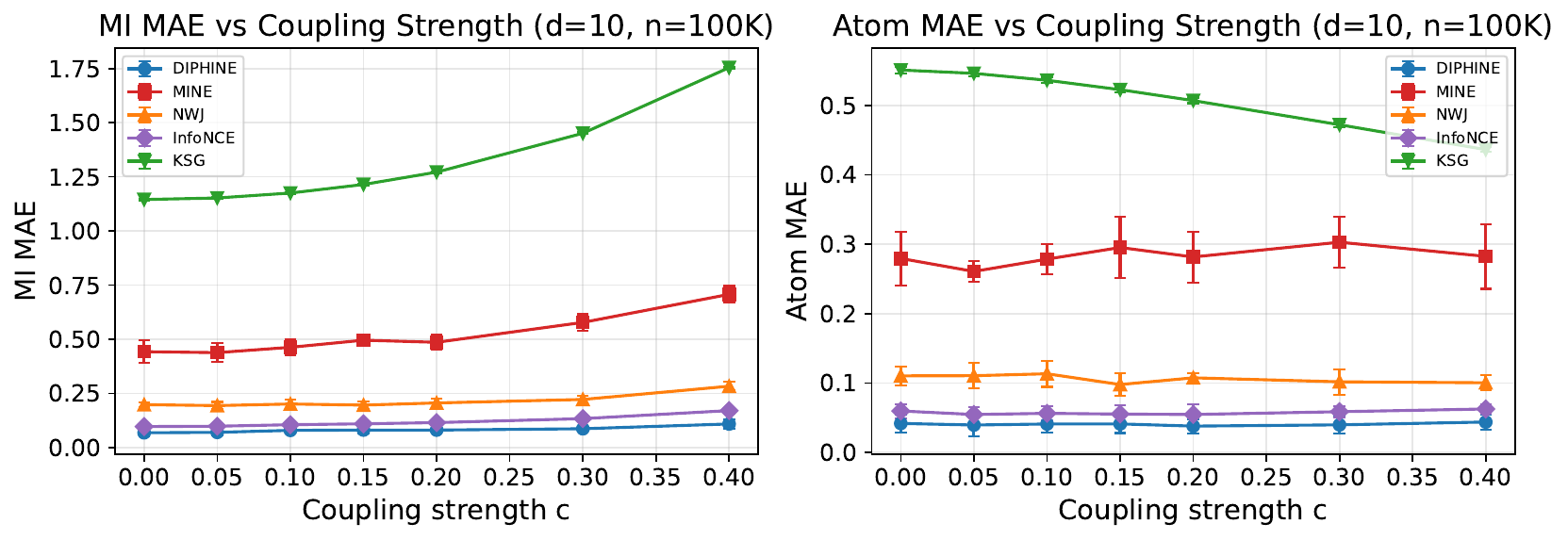}
    \caption{All methods at $d = 10$, $n = 100{,}000$. DIPHINE achieves the lowest atom MAE; MINE and KSG degrade catastrophically.}
    \label{fig:app_coupling_d10_all}
\end{figure}

For $d \in \{3, 5, 10\}$, we use the sparse coupled generator with coupling strength $c \in \{0, 0.05, 0.1, 0.15, 0.2, 0.3, 0.4\}$ and target spectral radius $0.85$, rescaling to ensure stationarity. All results are at $n = 100{,}000$. \Figref{fig:app_coupling_md} shows DIPHINE alone; Figures~\ref{fig:app_coupling_d3_all}--\ref{fig:app_coupling_d10_all} compare all methods at each dimension.

At $d = 3$, InfoNCE ($0.012$--$0.015$ MI MAE) and NWJ ($0.014$--$0.021$) remain competitive with or better than DIPHINE ($0.017$--$0.022$) on MI MAE. KSG degrades substantially ($0.073$--$0.119$), while MINE ($0.037$--$0.051$) is noticeably worse. At $d = 5$, InfoNCE ($0.018$--$0.024$) still achieves lower MI MAE than DIPHINE ($0.027$--$0.042$), and NWJ remains comparable. However, DIPHINE pulls ahead on atom MAE ($0.012$--$0.019$ vs.\ InfoNCE's $0.007$--$0.010$), reflecting its advantage in accurately translating MI estimates to atoms. KSG deteriorates further ($0.221$--$0.328$ MI MAE).

At $d = 10$, the picture changes decisively. DIPHINE achieves the lowest atom MAE ($0.038$--$0.044$), while InfoNCE ($0.055$--$0.063$), NWJ ($0.098$--$0.113$), MINE ($0.261$--$0.303$), and KSG ($0.436$--$0.551$) all substantially exceed it. On raw MI MAE, InfoNCE ($0.097$--$0.171$) remains the closest competitor to DIPHINE ($0.068$--$0.109$), but MINE ($0.438$--$0.707$) and KSG ($1.145$--$1.754$) collapse. These results show that DIPHINE's advantage is specifically on atom MAE at higher dimensions, where the shared score representation across all 9 MI terms provides a consistent benefit over independently trained estimators.
\newpage
\section{Additional real-data results}
\label{app:real_data}
\subsection{Full $\Phi$ID atom matrices}
\label{app:phiid_heatmaps}
\begin{figure}[H]
    \centering
    \includegraphics[width=0.85\textwidth]{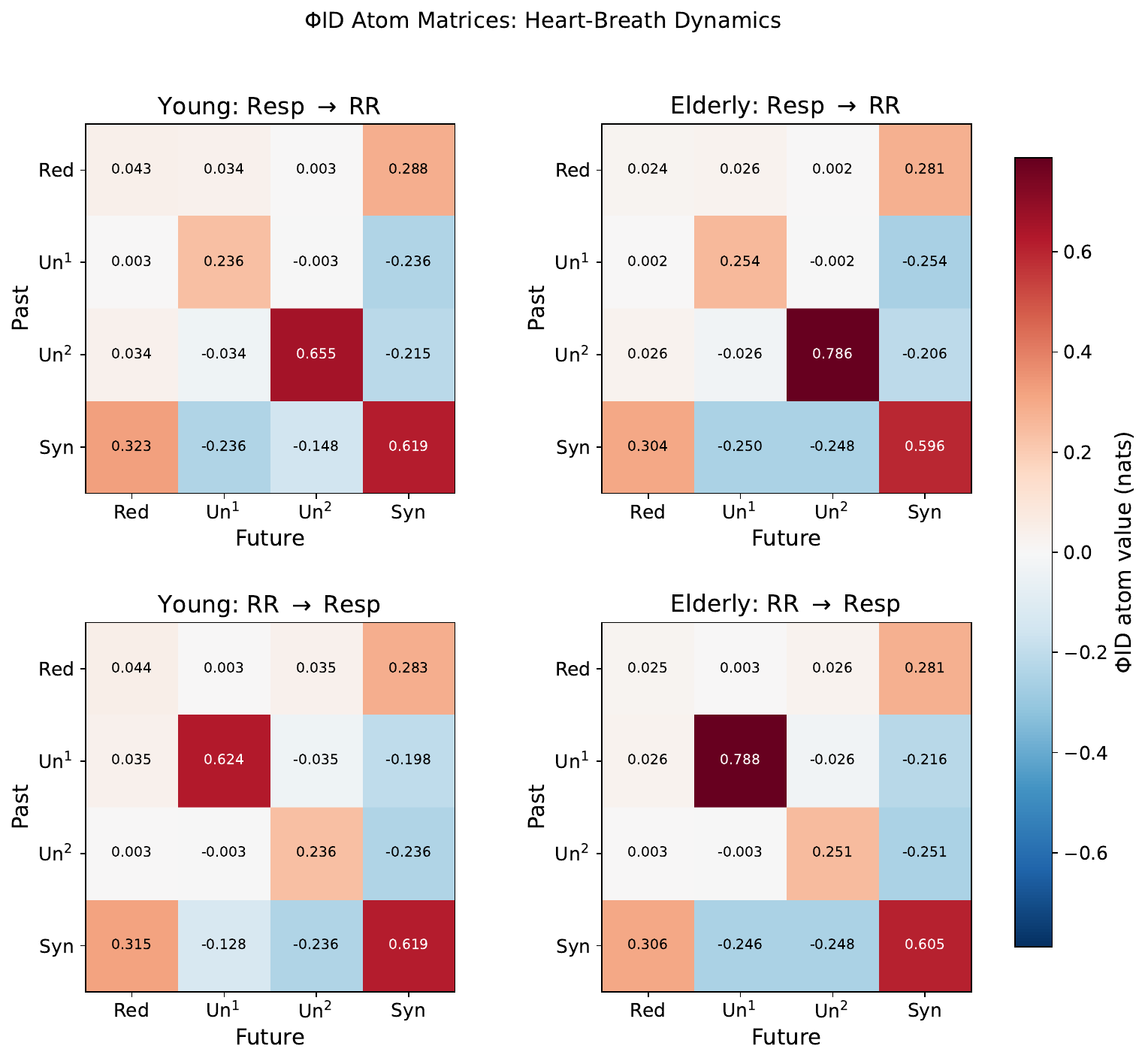}
    \caption{Full \gls{phiid} atom matrices for heart-breath dynamics. Top row: Resp$\to$RR. Bottom row: RR$\to$Resp. Left column: young. Right column: elderly. Values in nats.}
    \label{fig:app_phiid_heatmaps}
\end{figure}

\Figref{fig:app_phiid_heatmaps} shows the complete $4 \times 4$ atom matrices for both directions and both age groups. The largest atoms are the self-storage terms Un$_i \to$Un$_i$ and the synergistic integration Syn$\to$Syn. The overall structure is remarkably consistent between young and elderly subjects, with the primary difference being a quantitative increase in the self-storage atoms in the elderly group.
\subsection{Estimation stability across seeds}
\label{app:seed_stability}
\begin{figure}[H]
    \centering
    \includegraphics[width=0.6\textwidth]{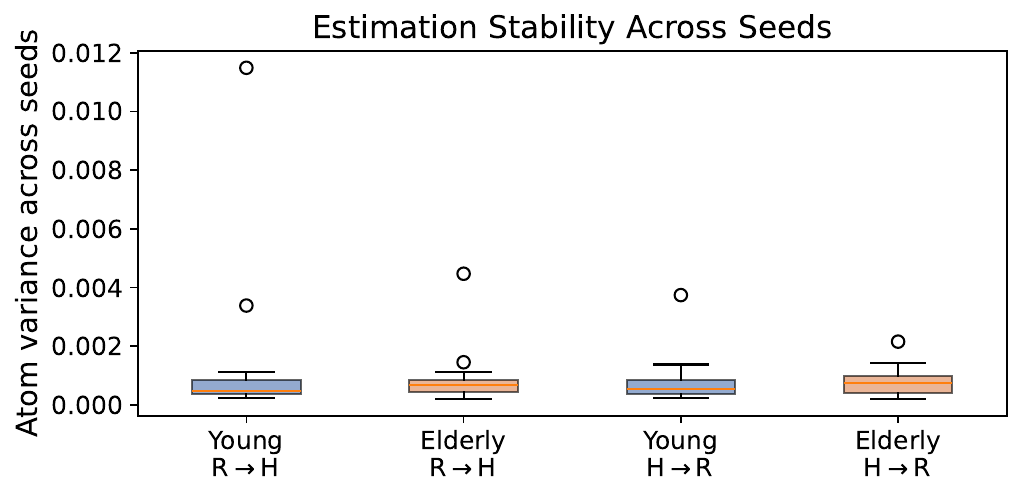}
    \caption{Mean cross-seed atom variance for each direction and age group in the Fantasia dataset. The median variance is below $0.001$ across all conditions.}
    \label{fig:app_seed_variance}
\end{figure}

To assess the reproducibility of the estimates, we examine the cross-seed variance of the 16 atom values across the 5 independent training runs for each subject. \Figref{fig:app_seed_variance} shows that the median cross-seed atom variance is below $0.001$ across all conditions, with occasional outliers that remain below $0.012$. This confirms that the estimation is stable and that the reported atom values are not dominated by stochastic variability in the training procedure.
\subsection{TE asymmetry histograms}
\label{app:te_asymmetry}
\begin{figure}[H]
    \centering
    \includegraphics[width=\textwidth]{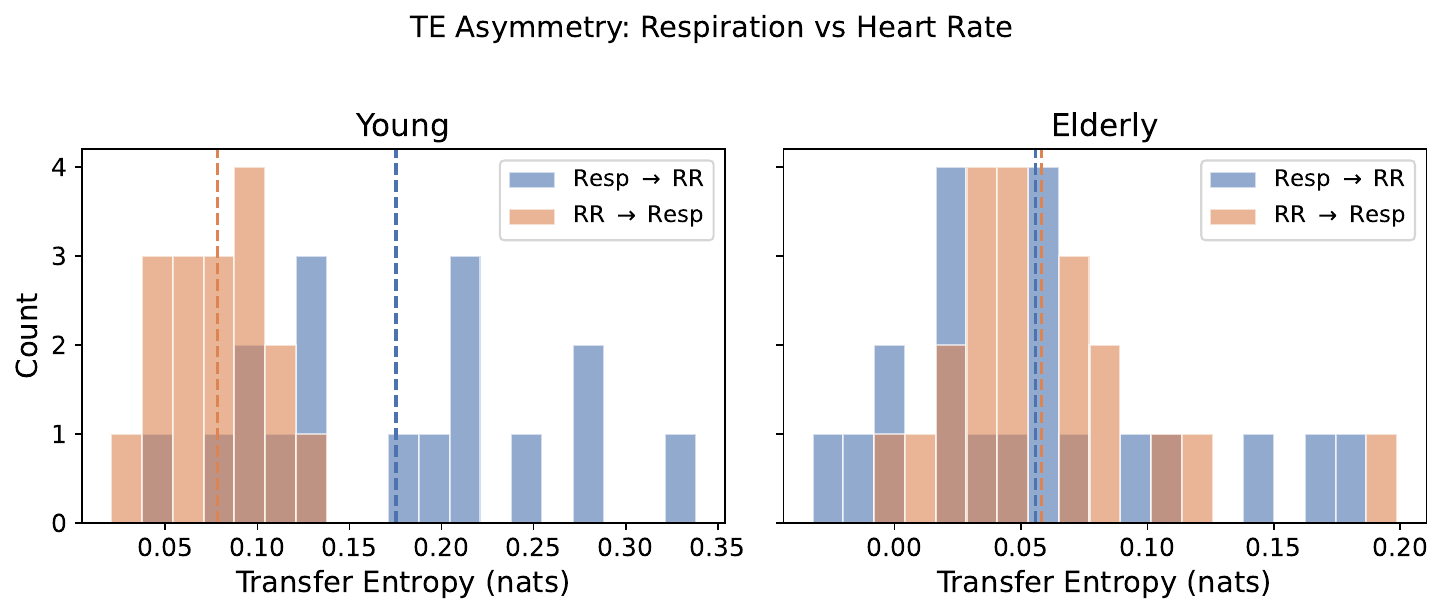}
    \caption{Distribution of \gls{te} estimates across subjects for both directions in young (left) and elderly (right) groups. Dashed lines indicate group means.}
    \label{fig:app_te_asymmetry}
\end{figure}

\Figref{fig:app_te_asymmetry} shows the distribution of \gls{te} estimates across individual subjects. In the young group, the Resp$\to$RR direction is consistently higher than RR$\to$Resp, with clear separation of the group means. In the elderly group, both directions are reduced and their distributions overlap substantially, consistent with the attenuation of directional asymmetry reported in \Secref{sec:real_data}.

\end{document}